\documentclass{article}

\usepackage{bm}
\usepackage{amsmath}
\usepackage{amssymb}
\usepackage{mathtools}
\usepackage{amsthm}
\usepackage{framed}
\usepackage[lighttt]{lmodern}

\usepackage{listings}
    \PassOptionsToPackage{numbers, compress}{natbib}

\usepackage[preprint]{neurips_2023}

\usepackage[utf8]{inputenc} 
\usepackage[T1]{fontenc}    
\usepackage{hyperref}       
\usepackage{url}            
\usepackage{booktabs}       
\usepackage{amsfonts}       
\usepackage{nicefrac}       
\usepackage{microtype}      
\usepackage{xcolor}         

\usepackage{enumitem}
\usepackage{subfigure}
\usepackage{multirow}
\usepackage{floatrow}
\usepackage{extarrows}
\usepackage{acronym}
\usepackage{xspace}
\usepackage{cleveref}
\usepackage{enumitem}
\usepackage[tableposition=top]{caption}
\newfloatcommand{capbtabbox}{table}[][\FBwidth]
\usepackage{wrapfig}
\usepackage{titletoc}
\usepackage{appendix}

\newcommand\DoToC{%
  \startcontents
  \printcontents{}{1}{\textbf{\Large Table of Contents}\vskip5pt\hrule\vskip3pt}
  \vskip3pt\hrule\vskip5pt
}

\makeatletter
\DeclareRobustCommand\onedot{\futurelet\@let@token\@onedot}
\def\@onedot{\ifx\@let@token.\else.\null\fi\xspace}
\def\eg{\emph{e.g}\onedot} 

\def\ie{\emph{i.e}\onedot}

\def\etc{\emph{etc}\onedot}

\makeatother

\acrodef{llm}[LLM]{Large Language Model}
\acrodef{kg}[KG]{knowledge graph}
\usepackage[lighttt]{lmodern}

\usepackage{listings}
\usepackage[T1]{fontenc}

\usepackage{bm}
\usepackage{amsmath}
\usepackage{amssymb}
\usepackage{mathtools}
\usepackage{amsthm}
\usepackage{framed}

\title{Large Language Models are In-Context \\ Semantic Reasoners rather than Symbolic Reasoners}

\author{%
 	Xiaojuan Tang\textsuperscript{1,3}\thanks{Equal contribution},
    Zilong Zheng\textsuperscript{3}$^*$,
	Jiaqi Li\textsuperscript{3},
    \\ \bf
	Fanxu Meng\textsuperscript{1,3},
	Song-Chun Zhu\textsuperscript{1,2,3},
	Yitao Liang\textsuperscript{1,3}\thanks{Corresponding authors},
    Muhan Zhang\textsuperscript{1,3}$^\dagger$
	\vspace{.5em} 
	\\ 
	\textsuperscript{1}Peking University,
    \textsuperscript{2}Tsinghua University
    \\
    \textsuperscript{3} National Key Laboratory of General Artificial Intelligence, BIGAI
    \\
}

\begin{document}

\maketitle


 
\begin{abstract}
    
     The emergent few-shot reasoning capabilities of Large Language Models (LLMs) have excited the natural language and machine learning community over recent years.
     Despite of numerous successful applications, the underlying mechanism of such in-context capabilities still remains unclear. In this work, we hypothesize that the learned \textit{semantics} of language tokens do the most heavy lifting during the reasoning process. Different from human's symbolic reasoning process, the semantic representations of LLMs could create strong connections among tokens, thus composing a superficial logical chain.
     To test our hypothesis,
     we decouple semantics from the language reasoning process and evaluate three kinds of reasoning abilities, \ie, deduction, induction and abduction. Our findings reveal that semantics play a vital role in LLMs' in-context reasoning---LLMs perform significantly better when semantics are consistent with commonsense but struggle to solve symbolic or counter-commonsense reasoning tasks by leveraging in-context new knowledge. 
     The surprising observations question whether modern LLMs have mastered the inductive, deductive and abductive reasoning abilities as in human intelligence, and motivate research on unveiling the magic existing within the black-box LLMs.
     On the whole, our analysis provides a novel perspective on the role of semantics in developing and evaluating language models' reasoning abilities. Our code is available at {\url{https://github.com/XiaojuanTang/ICSR}}.

     
\end{abstract}

\section{Introduction}



In recent years, \acp{llm} have achieved impressive performance on a variety of natural language tasks, including question answering, text summarization, machine translation, logic reasoning, \etc. These successes have been largely attributed to the emergent ability of \acp{llm} to utilize a ``zero-shot'' or ``few-shot'' learning approach without any gradient updates---a task description or a few examples are provided to guide their reasoning process~\cite{brown2020language,zhang2022automatic,bommasani2021opportunities,lu2021pretrained}. 
One typical example is the ``chain-of-thought (CoT)'' approach, involving reasoning demonstrations or a simple prompt such as ``Let's think step by step'' to perform complex reasoning tasks~\cite{wei2022chain,kojima2022large}.

Despite the powerful and versatile in-context learning ability of \acp{llm}, the underlying mechanisms by which they operate within a given context still remain unclear. Previous works investigate which aspects of the given examples contribute to the final task performance, including ground-truth labels and example ordering ~\cite{min2022rethinking,liu2021makes,zhao2021calibrate}. Another line of recent work has focused on explaining and leveraging the in-context learning (ICL) mechanism~\cite{schick2020exploiting,lu2021fantastically,akyurek2022learning,xie2021explanation}. However, the basic problem they have in common is that the in-context prompts they input are based on natural language queries to investigate the reasoning abilities of \acp{llm}. However, according to the Dual Process Theory~\cite{sloman1996empirical,gilovich2002heuristics}, humans usually use symbolic reasoning with System II to solve complex logical reasoning problems. To fill the research gap, we systematically study the in-context reasoning ability of \acp{llm} by decoupling the semantics from the language reasoning process. With extensive experiments, we aim to answer the following research question: \textit{Are \acp{llm} good in-context reasoners without semantics?}

In this work, we hypothesize that the learned semantics of language tokens play an important role in the reasoning process, creating strong connections among tokens which help to compose a superficial logical chain (shortcut) instead of really performing the formal reasoning process. To test our hypothesis, given symbolic knowledge (facts and rules), we test three kinds of reasoning abilities (\ie, deduction, induction, abduction) on a newly proposed synthetic dataset: Symbolic Tree dataset, which is composed of closed-world, noise-free, multi-hop symbolic reasoning data generated with logical rules. Besides, we also experiment with ProofWriter~\cite{tafjord2020proofwriter} task, containing questions whose answers require multi-hop reasoning. Our findings suggest that semantics indeed play a vital role in \acp{llm}' in-context reasoning: When semantics are consistent with commonsense, \acp{llm} perform fairly well; when semantics are decoupled or in the counter-commonsense context, \acp{llm} struggle to solve the reasoning tasks by leveraging in-context new knowledge. Moreover, we also study the memorization ability of LLMs to memorize new symbols and semantics information, allowing us to investigate the role of semantics on \acp{llm}' knowledge update ability.

To the best of our knowledge, this is the first study of the effect of semantics on \acp{llm}' in-context reasoning abilities. Our analysis underscores the importance of semantics in \acp{llm}' reasoning ability and questions whether modern \acp{llm} have mastered the formal reasoning abilities as in human intelligence. We hope our findings can provide a novel perspective on the role of semantics in \acp{llm}' in-context abilities, and inspire further research on unveiling the magic inside the black-box \acp{llm}.

\section{Related Works}


\textbf{Reasoning in \acp{llm}}~
Reasoning is a fundamental cognitive process involving logical inferences and conclusions based on given information. Developing models with strong reasoning capabilities has attracted increasing attention and many researches have been conducted on this topic since early days in the NLP domain~\cite{cooper1996fracas}. Since then, various benchmarks focusing on different aspects of reasoning have been proposed, including natural language inference (NLI)~\cite{Bowman_Angeli_Potts_Manning_2015, Williams_Nangia_Bowman_2017, Bernardy_Chatzikyriakidis_2020}, commonsense reasoning~\cite{Talmor_Herzig_Lourie_Berant_2018, Aggarwal_Mandowara_Agrawal_Khandelwal_Singla_Garg_2021}, multi-hop reasoning~\cite{Dua_Wang_Dasigi_Stanovsky_Singh_Gardner_2019, Chen_Zha_Chen_Xiong_Wang_Wang_2020} etc.
In recent years, there has been growing interests in studying the reasoning abilities of LLMs. Researchers have explored various approaches to enable LLMs to perform better on reasoning tasks. For example, ``chain-of-thought (CoT)''~\cite{wei2022chain,wang2022self} is proposed to facilitate models to generate a reasoning path that decomposes complex reasoning into multiple easier steps; LLMs are decent zero-shot reasoners by adding a simple prompt, ``Let’s think step by step'', to facilitate step-by-step thinking before giving the final answer~\cite{kojima2022large}. This significantly improves the performance on arithmetic~\cite{roy2016solving}, commonsense~\cite{Talmor_Herzig_Lourie_Berant_2018, geva2021did}, and symbolic reasoning~\cite{wei2022chain} benchmarks. However, despite their impressive performance on various reasoning benchmarks, all the tasks evaluated are rich in semantics. Thus it is unclear where the reasoning abilities of \acp{llm} come from. This motivates us to investigate \acp{llm}' reasoning abilities when semantics are decoupled.

\textbf{In-Context Learning}~
\acp{llm}' reasoning abilities are closely related to in-context learning (ICL). ICL refers to the ability of language models to adapt and learn from a few prompt examples during the inference process. In recent years, there has been a focus on exploring how to improve the performance of ICL. Specifically, some works select related demonstrations to the test instance using off-the-shelf unsupervised similarity metrics or train a prompt retriever to select examples~\cite{das-etal-2021-case, rubin2021learning, su2022selective}. Others incorporate task instructions or different task prompts~\cite{sanh2022multitask, wei2022finetuned}. Despite the empirical success, the underlying mechanisms of ICL still remain unclear. A few studies have shown that the performance of ICL usually varies with the choice of in-context demonstrations~\cite{liu2021makes, Lu_Bartolo_Moore_Riedel_Stenetorp_2021}. Specifically, the order of demonstrations may lead to large performance fluctuations~\cite{webson-pavlick-2022-prompt, zhao2021calibrate}. Recent works also explore the effect of ground-truth labels and question the necessity of ground-truth input-output mapping---using incorrect labels in the examples only marginally lowers the performance~\cite{Min_Lyu_Holtzman_Artetxe_Lewis_Hajishirzi_Zettlemoyer} and input-label correspondence plays a more important role in contextual demonstration~\cite{kim2022ground}. To further understand why in-context learning works, some work provides theoretical analysis that in-context learning can be formalized as Bayesian inference~\cite{xie2021explanation} or some instances of ICL can be understood as implicit implementation of known learning algorithms~\cite{urek_Schuurmans_Andreas_Ma_Zhou_2022}. However, the existing analyses of ICL are mainly based on natural language input with rich semantic information. We hypothesize that this might not be able to reflect their true level of reasoning abilities including deduction, induction and abduction. Thus, this paper aims to decouple semantics in \acp{llm}' in-context reasoning abilities.

\textbf{Symbolic Reasoning}~
Symbolic reasoning has long been studied in the field of artificial intelligence and cognitive science~\cite{mccarthy1981some,lavrac1994inductive,newell2007computer}. It involves manipulating symbols and applying logical rules to perform deduction~\cite{johnson1999deductive}, induction~\cite{lavrac1994inductive}, and abduction~\cite{kovacs2005abductive}.
\citet{boole1847mathematical} introduced Boolean algebra, which laid the foundation for symbolic logic and provided a formal system for logical reasoning.
\citet{mccarthy1960recursive} introduced LISP programming language and the concept of symbolic computation, which boosted the development of sophisticated AI programs that could represent and manipulate complex ideas and relationships.
\citet{fuhr2000probabilistic} introduced probabilistic Datalog, an extension of Datalog with probabilities, allowing for probabilistic reasoning in logic-based systems.
\citet{eiter2009answer} introduced answer set programming (ASP), a logic-based programming paradigm that combines logic programming and non-monotonic reasoning. ASP has been used for various reasoning tasks, including planning, knowledge representation, and constraint solving.
\citet{yi2018neural} proposed a neural-symbolic approach to visual question answering. It combines deep neural networks with symbolic rules to perform compositional and interpretable reasoning over visual and textual information.
\citet{shin2018improving} explore using LLM-based models for program synthesis. They present an approach that leverages inferred execution traces to guide the generation of correct programs.
\citet{lample2019deep} focus on applying LLM-based models to mathematical reasoning, proposing a framework that combines deep learning with symbolic mathematics to perform algebraic reasoning, equation solving, and theorem proving.
\citet{pallagani2022plansformer} use LLMs for automated planning---a branch of AI concerned with realizing action sequences (plans) to achieve certain goals, typically executed by intelligent agents, autonomous robots, and unmanned vehicles.

\section{Decoupling Semantics from In-Context Reasoning}

\subsection{Task Definitions}

\begin{figure}[t]
    \centering
    \includegraphics[width=\linewidth]{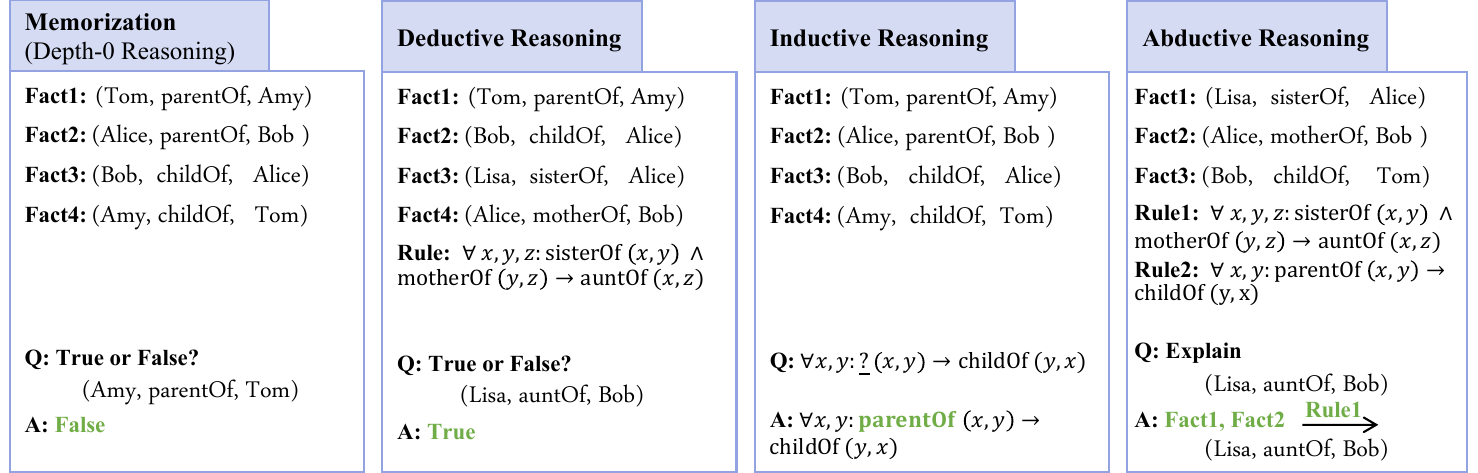}
    \caption{Task Definitions. \textbf{Memorization}: retrieving the predicted fact from in-context knowledge.
    \textbf{Deductive}: predicting the correctness of the predicted fact given rules and facts. \textbf{Inductive}: generating a rule based on multiple facts with similar patterns. \textbf{Abductive}: explaining the predicted fact based on given rules and facts. }
    \label{fig:tasks}
\end{figure}

To begin, we first introduce the definitions of reasoning and memorization mechanisms and provide task descriptions for each. Examples of the tasks are shown in Figure~\ref{fig:tasks}.

\textbf{Reasoning}~ In the field of psychology, reasoning refers to the process of using logical operations to draw conclusions or make inferences based on available information~\cite{wason1972psychology, Wason_2007, fagin2004reasoning, walton1990reasoning}. As an abstract notion, it encompasses a variety of aspects. Traditionally, we can classify it into three categories:
\begin{itemize}[leftmargin=.2in, topsep=0pt]
    \item \textit{Deductive reasoning} is a logical process in which a conclusion can be derived from given premises or principles, meaning predicting new facts based on existing facts and logical rules. For example, given the two facts (Lisa, sisterOf, Alice) and (Alice, motherOf, Bob) along with a logical rule $\forall x,y,z: \text{sisterOf}(x,y) \land \text{motherOf}(y,z) \rightarrow \text{auntOf}(x,z)$, the new fact (Lisa, auntOf, Bob) can be derived through deductive reasoning. The task is to predict the True/False of a predicted fact given facts and rules. The accuracy is the proportion of correct predictions.

    
    \item \textit{Inductive reasoning} involves making generalizations based on specific observations or evidence. In other words, a logical rule can be induced from given facts. For instance, 
    given a set of observations that person A is the parent of person B and person B is the child of person A, inductive reasoning is to conclude the logical rule $\forall x,y: \text{parentOf}(x,y) \rightarrow \text{childOf}(y,x)$. We perform the \textit{rule generation} task. Given multiple facts with similar patterns and a rule template, the goal is to induce a rule that entails these facts. We test the generated rules against the ground truth rules. If the generated rule matches the ground truth rule exactly, we regard the rule as correct; otherwise, we regard the rule as incorrect. The precision is the proportion of correct rules. More details of the rule template and the ground-truth rules are provided in Appendix~\ref{app:evaluation of induction}.

    \item \textit{Abductive reasoning} is a logical process of seeking a hypothesis that best fits or explains a set of observations. For example, given a lot of facts including (Lisa, sisterOf, Alice) and (Alice, motherOf, Bob), along with a set of logical rules including $\forall x,y,z: \text{sisterOf}(x,y) \land \text{motherOf}(y,z) \rightarrow \text{auntOf}(x,z)$, if we observe Lisa is Bob's aunt, one possible explanation is that Lisa is Alice's sister and Alice is Bob's mother. We use \textit{explanation generation} to evaluate the abductive reasoning ability. Given a \textit{theory} including facts and logical rules, the task is to select specific facts and a logical rule from the given theory to explain the \textit{observation}. The \textit{observation} is chosen from inferred facts. We use Proof Accuracy (PA) as an evaluation metric, \ie, the fraction of examples where the generated proof matches exactly any of the gold proofs.

\end{itemize}

\textbf{Memorization}~ 
Memory plays a crucial role in reasoning, as it involves storing the in-context or parametric knowledge necessary for the reasoning process. In some sense, memory can be considered as Depth=0 reasoning, where the question is a known fact. The reasoning task involves retrieving the fact itself from the stored knowledge. However, the specific impact of semantics on memorization has not been extensively explored. Thus, in addition to decoupling semantics from reasoning, we also try to study the impact of semantics on memorization. Specifically, we use a new dataset to fine-tune a language model and test its \textit{time}, \textit{efficiency} and \textit{forgetting ratio}: \textit{time} is the time cost of adding/updating facts, \textit{efficiency} is the filter MRR (the mean reciprocal of rank of the correct entity~\cite{bordes2013translating}) of the facts added/updated, and \textit{forgetting ratio} is the filter MRR of the facts that should not be updated. When evaluating whether a fact has been successfully added or updated, we query LLM with a question about the tail entity and rank the probability of the true tail against all entities. The better LLM remembers a triplet, the higher the MRR gets. 

\subsection{Evaluation Datasets}
Our goal is to decouple semantics from the in-context reasoning process and solely rely on the given (new) knowledge to perform reasoning tasks. To implement this, we use Symbolic Tree~\cite{hohenecker2020ontology} and ProofWriter~\cite{tafjord2020proofwriter} datasets, which contain both relevant and irrelevant facts and \acp{llm} need to infer the unknown facts after selecting relevant facts from memory.

The Symbolic Tree dataset is an artificially close-world and noise-free symbolic dataset generated with complex logical rules. The dataset consists of randomly sampled ``\textit{basic facts}'', which include gender information and ``parentOf'' relations among individuals. With the given logical rules, the dataset allows for reasoning about 28 different types of family relations, ranging from easy inferences (\eg, fatherhood), to more elaborate ones (\eg, a daughter of someone’s cousin). \textit{Facts} consist of \textit{basic facts} (in-context knowledge) and \textit{inferred facts} (what to reason). Note that Symbolic Tree is a close-world dataset, which means that any facts not presented in the dataset are assumed to be false. Thus, we construct the false facts by replacing the head entity or tail entity with a random entity as negative examples in \textit{inferred facts}. Considering the context window size limitation, we restrict each tree's depth to 5 to generate the dataset. We experiment with 10 sampled Symbolic Trees; each has 30 kinds of relations (28 inferred relations, gender and parentOf relation), 26 entities, about 35 basic facts, 300 inferred facts and 300 false ones.

To decouple the semantics within the dataset, we replace the relation names (such as ``parent'') with hand-crafted symbols (\eg, ``r1'', ``r2'', ...), so that LLMs cannot leverage the semantics of the predicates in reasoning but must resort to the given new knowledge (presented as in-context facts and rules). We also experiment with replacing entity names (such as ``Alice'') with ``e1'', ``e2'', ..., but find that it has little impact on performance (more details are provided in Appendix~\ref{entity labels}). During the symbol generation process, we also try to randomly sample some letters as relation names (\eg, ``lnqgv'' instead of ``r1''), but we observe that \acp{llm} struggle to understand garbled characters, which may negatively affect performance (further discussion is provided in Appendix~\ref{garbled symbols}).

\begin{wrapfigure}{r}{0.5\textwidth}
  \centering
    \includegraphics[width=\linewidth]{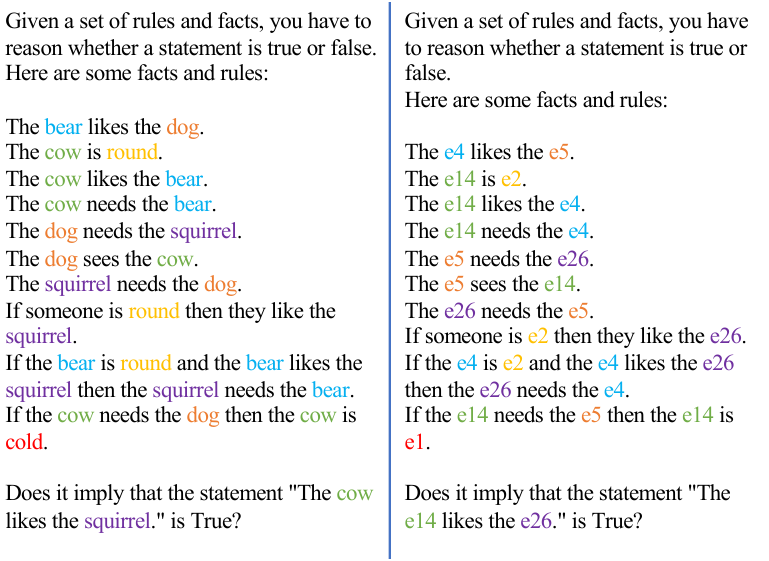}
    \caption{Decoupling semantics from the ProofWriter task. In the original ProofWriter task, entities are represented by their names (left). However, in our decoupled setting, we replace the entity names with unique entity IDs (right).}
    \label{fig:proofwriter}
\end{wrapfigure}

ProofWriter~\cite{tafjord2020proofwriter} tasks provide artificial facts and rules expressed in natural language. For our experiments, we use a subset of the ProofWriter Open World Assumption (OWA) dataset with a depth of 1, 2, 3 and 5 (there is no depth 4 task), which contains many small rulebases of facts and rules, expressed in English and do not exist in \acp{llm}' knowledge base. Each rulebase has a set of questions (English statements) that can be proven true, false or ``Unknown''. Note that if we want to prove something Unknown, it is necessary to enumerate all possible facts and check their true/false. Thus, we remove all the Unknowns and replace the subject and object with entity IDs. This dataset is simpler than Symbolic Tree. Considering most of the predicates in the sentences are unmeaningful verbs like ``is'' and ``can'', we only replace the entities with entity IDs to decouple semantics. Take ``Anne is kind.'' as an example. We substitute subject (Anne) and object (kind) with ``e1'' and ``e2'', respectively, resulting in ``e1 is e2''. Figure~\ref{fig:proofwriter} provides an illustrative example.




\setlength{\intextsep}{0pt plus 2pt}

\section{Experiment}

\textbf{Models Selected for Evaluation}~
We primarily evaluate the performance of ChatGPT, GPT-4 and LLaMA. \textbf{ChatGPT} and \textbf{GPT-4} are advanced AI models developed by OpenAI and have demonstrated strong reasoning abilities across various tasks and benchmarks. \textbf{LLaMA} is an open-source large language model developed by Meta AI, with number of parameters ranging from 7B to 65B. Due to computational resource constraints, we could only fine-tune the LLaMA-7B version, which is used in our memorization test. As a reference, when comparing the reasoning abilities of LLMs, we also use some \textbf{logic-based} symbolic methods to conduct experiments as the baseline. To compare memorization, we use a popular graph database \textbf{Neo4j}~\cite{carro2014nosql} as the baseline. To ensure a relatively fair comparison, we configure Neo4j with a pre-stored knowledge base that has comparable disk space size to LLaMA.
 More introduction of Neo4j is represented in Appendix~\ref{app:neo4j}.

\textbf{Evaluation Setup}~
For reasoning, we use Symbolic Tree and ProofWriter Depth-1 as evaluation data. We refer to the raw data, where semantics are retained, as \textit{Semantics}. When semantics are decoupled using symbols, we refer to it as \textit{Symbols}. For the Symbolic Tree dataset, we experiment with 10 sampled trees and report the average results, where facts and rules can be represented as logical language and natural language text as the input of \acp{llm}.  For example, the fact ``motherOf(Alice, Bob)'' can be represented as ``Alice is Bob's mother''; the rule ``$\forall x,y: \text{parentOf}(x,y) \rightarrow \text{childOf}(y,x)$'' can be represented as ``If x is parent of y, then y is child of x.''. Through numerous trials, we find that for the \textit{Symbols} setting, \acp{llm} tend to perform better when using logic language representations. Conversely, for the \textit{Semantics} setting, \acp{llm} tend to perform better when using natural language text. We select the representation that yields better performance in \acp{llm}' reasoning. Additional results are presented in Appendix~\ref{app:results of two repre}. We consider zero-shot, zero-shot CoT, few-shot CoT and zero-plus-few-shot-CoT as baselines. To generate explanations for few-shot CoT experiments, for deductive reasoning, we use zero-shot CoT (\ie, Let's think step by step) to generate explanations given the random questions; for abductive reasoning, we randomly select five examples and manually design their demonstrations. 
We provide all prompts and CoT demonstrations in Appendix~\ref{prompts}.

For memorization, we randomly selected 1,258 triplets from four sampled Symbolic Trees to fine-tune the LLaMA. We then select the triplets in the first two trees and update their tail entities. The rest two trees are used to evaluate the  forgetting ratio.
Note that when updating, we simply flip the tails to random false facts and do not consider the consistency of the knowledge base. More illustrations of consistency are in Appendix~\ref{updating}. We still use \textit{Symbols} and \textit{Semantics} to denote different experiment settings. Both settings ensure that the new information provided does not overlap with the old knowledge base of LLMs, avoiding any ambiguation problems and eliminating the influence of pre-existing knowledge on the memorization task. When testing, we follow the prompting of~\citet{alpaca}, using the head entity and relation as instructions and providing all candidate tails as input. The detailed prompts are contained in Appendix~\ref{prompts}.

\textbf{Implementation Details}~
For ChatGPT and GPT-4, we use the chat completion API provided by OpenAI. We use a temperature of zero to generate output. Additionally, we set the frequency penalty to zero and top p to 1, which are the default values for these APIs. 

For LLaMA-7B, we utilized 4 A100 80G GPUs with batch size 64 for finetuning. The training process involved 100 epochs, employing a cosine learning rate schedule with an initial learning rate of 2e-5. We run these experiments three times and recorded their mean MRR and standard deviations. Please refer to Appendix~\ref{implementation of memorization} for more details.

For logic-based symbolic baseline,
in the deductive reasoning setting, it enumerates paths between head h and tail t and uses activated rules to infer the answer; 
For inductive reasoning, we adopt AMIE+~\cite{galarraga2015fast}, which first enumerates possible rules and then learns a scalar weight for each rule to encode its quality. 
For abductive reasoning, we locate the logical rule that reason about the relation of the fact and find all paths connecting the head and tail that can activate the rule. These path facts, along with the logical rule, serve as the explanations. 

\subsection{Semantics Matter in \acp{llm}' memorizing}\label{memorizing and updating} 

We first test the memorization ability of \acp{llm} when new knowledge are presented in semantics/symbols forms. The results are reported in Table~\ref{memory}. 

\textbf{Results}~
From Table \ref{table:llm_kg}, the \textit{efficiency} of adding and updating semantic knowledge is higher compared to symbolic knowledge. This suggests that semantic knowledge is easier for \acp{llm} to memorize than symbolic knowledge, similar to human's memory capabilities (memorizing symbols is generally more challenging than memorizing words with semantic meanings). However, we also find that the \textit{forgetting ratio} of the semantic setting is higher than the symbolic setting. This could be attributed to the fact that semantic knowledge has stronger correlation with each other than symbolic knowledge in \acp{llm}, which means that updating some knowledge can have a more significant influence on other knowledge in the system.

In the comparison between \acp{llm} and the graph-based system Neo4j, we can see that knowledge update using Neo4j achieves 100\% accuracy when inserting new triplets or editing existing triplets, regardless of whether the knowledge is symbolic or semantic. As expected, since the added or updated knowledge does not overlap with the existing knowledge base, there is no further influence on the existing knowledge in the database. This highlights one advantage of using a non-parametric knowledge base. Additionally, compared to the computational cost of fine-tuning \acp{llm}, updating knowledge in a graph database with optimized storage mechanisms is significantly faster. This inspires that combining \acp{llm} with a non-parametric knowledge base can offer a more practical and efficient approach for real-world applications.





\begin{table}[t]
\caption{Memorization abilities: LLaMA-7B and Neo4j. MRR are in \%.}\label{memory}
\resizebox{\linewidth}{!}{
\begin{tabular}{cc|cccc}
\toprule
   \textbf{Method}& \textbf{Category} & Adding Efficiency (MRR)    & Updating Efficiency (MRR)     & Forgetting (MRR$\downarrow$) &Time/1k triplets\\ 
    \midrule
\multirow{2}{*}{\textbf{LLaMA-7B}}&\textbf{Semantics}         & $ 50.375 \pm 1.27 $          &  $51.34 \pm 0.55 $        & $7.02 \pm 1.55 $ 	&41.5 min\\
&\textbf{Symbols}                                       & $48.91 \pm 4.3 $          &  $40.74 \pm 2.26 $         & $2.2 \pm 0.99 $ & 41.5 min\\
\midrule
\multirow{2}{*}{\textbf{Neo4j}} &\textbf{Semantics} & 100  & 100  & 0 & 19s \\ 
 &\textbf{Symbols} & 100  & 100  & 0 & 19s \\ 
\toprule
\end{tabular}}
\label{table:llm_kg}
\end{table}

\begin{table}\small
  \caption{The reasoning results of Symbolic Tree. Results are in \%.}
  \label{tab:reasoning}
  \centering
  \begin{tabular}{ccc|ccc}
    \toprule
    \textbf{Category} &
    \textbf{Model}  &
    \textbf{Baseline} & deduction & induction & abduction \\
    \midrule
    
    \multirow{7}{*}{\textbf{\shortstack{Symbols}}} 
    & \multirow{4}{*}
    {\textbf{\shortstack{ChatGPT}}} 
      & Zero-Shot     &52.6 &6.10 &1.50\\
   
   & & Zero-Shot-CoT  &55.7  &7.86 &4.90    \\
    & & Few-Shot-CoT  &54.8  &-  &18.2 \\
    & & Zero-Plus-Few-Shot-CoT &55.7 &- &- \\
    & \multirow{3}{*}
    {\textbf{\shortstack{GPT-4}}} 
      & Zero-Shot   & 68.8 &9.28 &25.0\\
   
   & & Zero-Shot-CoT   &71.1 &8.93 &31.2 \\
    & & Few-Shot-CoT & 67.6 &- &44.2\\

\midrule
 \multirow{7}{*}{\textbf{\shortstack{Semantics}}} &
    \multirow{4}{*}
    {\textbf{\shortstack{ChatGPT}}} 
      & Zero-Shot     &66.1  &36.4 &2.94\\
     & & Zero-Shot-CoT &65.5  &32.2  &3.40 \\
      
   & & Few-Shot-CoT     &67.1  &- &21.8\\
  
    & & Zero-Plus-Few-Shot-CoT &67.2 &- &-   \\
   
   & \multirow{3}{*}
    {\textbf{\shortstack{GPT-4}}} 
      & Zero-Shot    &79.2 &52.5 &27.3 \\
   
   & & Zero-Shot-CoT    &86.2 &53.9 &33.4\\
    & & Few-Shot-CoT  & 91.1 &- &69.2 \\

    \midrule
    & {\textbf{\shortstack{Random}}} &- &50.1 &3.57 &-\\
    \midrule 
    & \textbf{Logic-based} &- &100 &57.1 &100 \\
    \toprule

  \end{tabular}
\end{table}


\begin{table}\small
  \caption{The deduction results of ProofWriter tasks (ChatGPT). Results are in \%.}
  \label{tab:reasoning-of-proofwriter}
  \centering
  \begin{tabular}{cc|cccc}
    \toprule
    \textbf{Category} &
    \textbf{Baseline} & depth-1 & depth-2 & depth-3 &depth-5 \\
    \midrule
    
    \multirow{3}{*}{\textbf{\shortstack{Symbols}}} 

      & Zero-Shot  &69.1 &62.3 &59.4 &52.8  \\
   
   & Zero-Shot-CoT &56.2  &49.4 &45.2 &38.6 \\
  & Few-Shot-CoT   &65.8 &58.1 &- &-\\

\midrule
 \multirow{3}{*}{\textbf{\shortstack{Semantics}}} &
       Zero-Shot      &69.0 &63.5 &60.3 &51.4\\
      & Zero-Shot-CoT &51.5 &45.8 &40.3 &30.9\\
      
    & Few-Shot-CoT  &62.5 &56.7 &- &- \\

\toprule

  \end{tabular}
\end{table}





\subsection{Semantics Play a Vital Role in LLMs' Reasoning}

In this section, we evaluate the impact of decoupling semantics from LLMs' in-context reasoning. In Table~\ref{tab:reasoning}, we present the results of deductive, inductive, and abductive reasoning tasks on the Symbolic Tree datasets.

\textbf{Results}~
From Table~\ref{tab:reasoning}, we observe that in both deductive and inductive reasoning scenarios, \textit{Semantics} setting significantly outperforms \textit{Symbols} setting. Notably, in the inductive experiments, \textit{Semantics} achieves approximately 30\% higher absolute accuracy compared to \textit{Symbols} setting. This indicates that preserving rich semantics in the reasoning process leads to better performance for \acp{llm}.

Although in the abductive reasoning scenario, \textit{Symbols} setting achieves comparable results to \textit{Semantics} setting, it is important to note that the longer in-context knowledge in abductive reasoning tasks can increase the difficulty of selecting relevant information and reasoning from memory. To further investigate this, additional experiments on a smaller Symbolic Tree dataset are conducted, and the results in Appendix~\ref{app:shorter context} confirm that the \textit{Semantics} setting still outperforms the \textit{Symbols} setting. This reinforces the hypothesis that preserving semantics enhances the reasoning capabilities of \acp{llm}.

Despite the improved in-context reasoning performance of \acp{llm} with rich semantics, when compared to logic-based symbolic methods, \acp{llm} still exhibit inferior performance in all reasoning tasks. This suggests that while \acp{llm} possess a broad knowledge base and strong language understanding, symbolic reasoning is not their primary strength compared to methods specifically designed for symbolic reasoning. This also suggests the potential of future neural-symbolic AI systems.





\subsection{More Fine-grained Analysis about Semantics}
The aforementioned experiments offer initial evidence highlighting the significance of semantics in the reasoning of \acp{llm}. To further investigate this observation, we examine the influence of commonsense knowledge stored within \acp{llm} on their semantic reasoning performance. Specifically, we explore three aspects: First, we investigate the impact of removing given logical rules (in deduction) and facts (in induction), where LLMs have to rely solely on the prior commonsense knowledge stored within the parameters to infer the answers. This analysis allows us to assess the extent to which \acp{llm} can leverage their internal knowledge to reason effectively without explicit in-context knowledge. Second, we retain the semantics of the datasets but introduce counter-commonsense logical rules. This requires \acp{llm} to leverage in-context new knowledge and navigate the reasoning process by strictly adhering to the new information conflicting with the old knowledge. We implement it by shuffling relations as new relation labels to construct a new counter-commonsense dataset. For instance, we replace ``motherOf'' with ``sisterOf'', ``parentOf'' with ``brotherOf'', and ``female'' with ``male''. Consequently, for a rule such as $\forall x,y:\text{parentOf}(x,y) \land \text{female}(x) \rightarrow \text{motherOf}(x,y)$, we obtain $\forall x,y: \text{brotherOf}(x,y) \land \text{male}(x) \rightarrow \text{sisterOf}(x,y)$. Thirdly, we use a subset of the ProofWriter OWA datasets for depths 0, 1, 2, 3 and 5, which contains synthetic facts and rules despite written in natural language but irrelevant to commonsense (see Figure~\ref{fig:proofwriter}). These investigations allow us to gain deeper insights into the effect of semantics on the reasoning capabilities of \acp{llm}. 

\textbf{When semantics are consistent with commonsense}~
As shown in \Cref{tab:counter-commonsense}, in the deductive reasoning experiment, \textit{Removing rules/facts} achieves comparable results to \textit{Semantics}; in the inductive reasoning experiment, \textit{Removing rules/facts} outperforms \textit{Symbols}, achieving 35.7\% in GPT-4. These findings suggest that \acp{llm} can perform deductive reasoning comparably by leveraging their stored commonsense knowledge without using the provided semantic knowledge, and providing symbolic instead of semantic knowledge in induction might even hurt the performance. Besides, GPT-4 significantly outperforms ChatGPT across all evaluation settings, suggesting that the commonsense knowledge stored within GPT-4 is more extensive and allows stronger reasoning capabilities.

\begin{wraptable}{r}{0.6\textwidth}
  \centering
  \captionof{table}{Semantics, removing rules/facts and counter-commonsense reasoning experiments (ChatGPT and GPT-4). Results are in \%.}\label{tab:counter-commonsense}
\resizebox{\linewidth}{!}{%

\begin{tabular}{c|cc|cc}
\toprule                & \multicolumn{2}{c|}{deductive (Few-Shot-CoT)}    & \multicolumn{2}{c}{inductive (Zero-Shot-CoT) }   \\  
& ChatGPT & GPT-4 & ChatGPT & GPT-4 \\
\midrule
    Semantics       &71.8 &  90.0    &25.0  & 53.6	 \\
    Symbols        &53.7  &  67.6   &7.14 &   21.4   \\
    Remove R/F & 70.1   &   90.4  &7.14      &  35.7\\ 
    Counter-CS  & 48.9   & 73.4 &7.14 & 17.8
\\

\toprule
\end{tabular}
}
\end{wraptable}

\textbf{When semantics are not consistent with commonsense}~
To investigate the impact of semantics that are not consistent with commonsense, we introduce counter-commonsense (Counter-CS) scenarios, which is also shown in \cref{tab:counter-commonsense}. In comparison to \textit{Semantics} and \textit{Symbols}, we find that \textit{Counter-Commonsense} performs worse than \textit{Semantics}, even \textit{Symbols}. These findings suggest that when the in-context new knowledge conflicts with commonsense, \acp{llm} struggle to accurately reason and predict.



\textbf{When semantics are irrelevant to commonsense}~We use the ProofWriter tasks to test whether unmeaningful semantics are still useful. The results are shown in \cref{tab:reasoning-of-proofwriter}. The \textit{Symbols} setting performs comparably to the \textit{Semantics} setting in the zero-shot setting, suggesting that when semantics are irrelevant to commonsense, they have little effect on the reasoning abilities of \acp{llm}. In other words, when the task does not require deep semantic understanding or relies minimally on commonsense knowledge, the presence or absence of semantics does not significantly impact the performance of \acp{llm}. However, in the CoT settings, we observe that \textit{Semantics} is significantly worse than \textit{Symbols}. This might be because step-by-step reasoning magnifies the disturbing effect brought by weird semantics such as ``The squirrel needs the dog''.
Additionally, we observe that the CoT settings even perform worse than the zero-shot setting, with a higher frequency of the answer ``Cannot be determined.''. Similar phenomenons are also observed in \cref{tab:reasoning}, indicating that CoT may not be always helpful for reasoning tasks with in-context new knowledge.

\subsection{More analysis and discussions}

\textbf{(1) Deduction outperforms other reasoning types:}~
Across both the \textit{Symbols} and \textit{Semantics} settings, \acp{llm} demonstrate better performance in deduction compared to induction and abduction. This observation may be attributed to the transformer-decoder architecture used by the GPT series, which predicts the next token based solely on the preceding tokens. This architecture aligns well with the deductive reasoning process, where conclusions are derived from given premises or principles. Intuitively, induction/abduction is also harder than deduction for human.

\textbf{(2) Shorter in-context knowledge enhances reasoning performance}~
To examine the influence of context length on reasoning, we conducted an abductive reasoning experiment using a smaller Symbolic Tree, containing approximately 12 entities and 100 facts. The results, provided in Appendix~\ref{app:shorter context}, show that abductive reasoning with a shorter context leads to better performance compared to a longer context. Besides, we also conduct deduction and induction experiments where LLMs are directly provided with the relevant facts related to the predicted fact or the predicted rule. The results are presented in Appendix~\ref{app:selection}. This finding suggests that \acp{llm} struggle with processing excessively long in-context information, particularly in reasoning tasks.
The length of the context influences reasoning performance, as shorter contexts make it easier to select relevant and useful information while minimizing the impact of unrelated content.



\textbf{(3) Effectiveness of commonsense expressed in natural language:}~
We explore the representation of knowledge in natural language and logic language forms in our experiments. The results, presented in Appendix~\ref{app:results of two repre}, indicate that for tasks involving semantics, natural language descriptions are more effective than logical language representations. Conversely, for symbolic and counter-commonsense tasks, logic language performs better. This observation suggests that natural language representations better stimulate the semantic understanding capabilities of \acp{llm}, while logical language representations are more conducive to symbolic reasoning.


\textbf{(4) Zero-shot capabilities are approaching zero-shot-CoT capabilities:}~
In \textit{Symbols} setting, comparing zero-shot with zero-shot-CoT across deduction, induction, and abduction evaluations, we observe that zero-shot-CoT only marginally improves the performance compared to zero-shot learning. This finding suggests that the zero-shot capabilities of current \acp{llm} are approaching their zero-shot-CoT learning abilities. One plausible explanation is that ChatGPT has already been trained on similar tasks with CoT and has memorized the instructions. Consequently, it implicitly follows these instructions when applied to the same queries, even without explicit CoT guidance~\cite{chen2023you}.

\textbf{(5) Zero-shot as System-I, Zero-shot-CoT as System-II}~
For \textit{Symbols} setting, comparing zero-shot with zero-shot-CoT across deduction, induction, abduction, we observe that 
zero-shot-CoT improves the performance compared to zero-shot learning. In contrast, for \textit{Semantics} setting, zero-shot-CoT underperforms zero-shot. This distinction can be attributed to the impressive zero-shot capabilities of \acp{llm} in various System-I tasks, highlighted by \citet{radford2019language}. Tasks that demand a profound comprehension of rich semantic information lead \acp{llm} to rely on intuition and employ shortcuts to predict answers. On the other hand, for symbolic reasoning tasks, zero-shot-CoT learning outperforms zero-shot learning, aligning with the findings of \citet{kojima2022large} who demonstrated the facilitative role of CoT in activating System-II abilities.


\textbf{(5) Utilizing internal knowledge outperforms external in-context knowledge:}~
To explore the ability of \acp{llm} to utilize internal and external knowledge, we conduct an additional experiment where we provide \acp{llm} with only the relevant facts related to the predicted fact. We compare the performance of \textit{Removing rules} (leveraging internal knowledge) with \textit{Semantics} (providing external logical rules). Surprisingly, we find that \textit{Removing rules} performed better than \textit{Semantics}. This suggests that \acp{llm} possess the necessary internal knowledge to support answering questions and reasoning tasks, and leveraging this internal knowledge is more effective for reasoning than relying on external logical rules. Detailed results and case studies can be found in Appendix~\ref{app:selection-deduction}.




\subsection{Commonsense Reasoning}\label{commonsense reasoning}
Commonsense reasoning refers to the ability to make judgments and draw conclusions based on understanding basic world knowledge and everyday experiences. For example, ``\textit{water is wet and can put out a fire}'' is an example of commonsense. By considering the properties of water and its ability to put out flames, we can deduce that ``pouring water on a kitchen fire'' would be an effective solution. This section explores the commonsense reasoning abilities of \acp{llm}. Although commonsense reasoning cannot decouple semantics, we include the comparisons for a thorough evaluation of LLMs' reasoning ability.

\textbf{Evaluation Setup}~
We use CommonsenseQA and OpenBookQA~\cite{mihaylov2018can} to test commonsense reasoning. We use ChatGPT, GPT-4 and GreaseLM~\cite{zhang2022greaselm} as baselines. More experimental setup is in Appendix~\ref{GreaseLM}.



\begin{table}[t]
\caption{The commonsense reasoning results of CSQA and OpenBookQA. Results are in \%.}\label{tab:commonsense}
\begin{tabular}{cc|ccc}
\toprule
Model                    & Baseline                                                             & CSQA-Dev                      & CSQA-Test              & OpenBookQA-Test       \\ \midrule
GreaseLM                       & -                                                     & 78.5                     & 74.2                     & 84.8  \\ \midrule
\multirow{6}{*}{ChatGPT} & Zero-Shot                                                            & 71.9                     & 67.8                     & 70.8 \\
                         & \begin{tabular}[c]{@{}c@{}}Few-Shot (predict-explain)\end{tabular} & 58.9                     & 57.6                     & 74.6 \\
                         & \begin{tabular}[c]{@{}c@{}}Few-Shot (explain-predict)\end{tabular} & 46.4                     & 44.3                     & 70 \\
                         & Few-Shot (3)                                                         & 68.3                     & 66.8                     & 75 \\
                         & Few-Shot (8)                                                         & 70.4                     & 68.8                     & 74 \\
                         & Few-Shot (16)                                                        & 70.5                     & 69.4                     & 73.8 \\ 
                        \midrule
\multirow{2}{*}{GPT-4} & Zero-Shot                                                  & 82.1      &  79.4   & 91.8 	\\
                         & Few-Shot                                                 &  85.3    & 82.5    &  86.2 \\ \toprule
\end{tabular}
\end{table}

\textbf{Results}
As shown in Table~\ref{tab:commonsense}, GreaseLM demonstrates remarkable performance in commonsense reasoning tasks, comparable to human performance of 88\% as indicated by a public leaderboard. In addition, we conduct experiments employing different few-shot prompts to further investigate its capabilities. The results indicate that few-shot prompts only yield marginal improvements compared to zero-shot learning. This phenomenon can be attributed to the Robust Human Language Filter (RHLF) of ChatGPT, which empowers the models to better understand the meaning and context of tasks, enabling them to perform well even in zero-shot settings.

However, when we examine the few-shot (predict-explain or explain-predict) results of CSQA, which involves CoT explanations, the performance falls short. This is probably because the questions of CSQA rely more on intuition (System-I) but less on explicit reasoning (System-II). GPT-4 exhibits significant improvements in its reasoning ability on both datasets. It achieves a success accuracy of over 80\% on CSQA and approaches an impressive 90\% on OpenBookQA. These results demonstrate the potential for further optimization of LLMs in commonsense reasoning through training on higher-quality corpora, using larger models, and using more iterations of training.

\section{Conclusion and Discussion}

Our paper presents the first comprehensive investigation of the role of semantics in \acp{llm}' in-context reasoning abilities by decoupling semantics from in-context prompts. Experimental results suggest that: When semantics are consistent with commonsense, \acp{llm} perform fairly well; when semantics are decoupled or counter-commonsense, \acp{llm} struggle to solve the reasoning tasks by leveraging in-context new knowledge. These findings reveal the importance of semantics in \acp{llm}' reasoning abilities and inspire further research on unveiling the magic existing within the black-box \acp{llm}. In light of the findings identified in our analysis, we point out several potential future directions for the development of large foundation models:

\textbf{More complex symbolic reasoning benchmark:}~ To improve \acp{llm}' in-context symbolic reasoning abilities, developing new datasets with decoupled semantics and more complex reasoning tasks is necessary. These benchmarks should challenge \acp{llm} with diverse and intricate symbolic knowledge.

\textbf{Combination with external non-parametric knowledge base:}~As our experimental results show, the memorization abilities of \acp{llm} are not comparable to existing graph-based methods. This motivates integrating \acp{llm} with external non-parametric knowledge bases, such as graph databases, to enhance their knowledge insertion and updating. This hybrid approach can leverage the strengths of \acp{llm}' language understanding and the comprehensive, accurate and up-to-date knowledge stored in non-parametric sources.

\textbf{Improving the ability of processing in-context knowledge:}~ More robust and strong abilities to process and memorize in-context knowledge is crucial to perform complex in-context reasoning tasks. Further research is needed to improve \acp{llm}' capabilities in processing and leveraging in-context knowledge. This includes developing mechanisms to better encode and retrieve relevant information from the in-context knowledge, in order to enable more effective reasoning.


\clearpage

{

\small
\bibliographystyle{unsrtnat}
\bibliography{ref}

}
\newpage













\title{\textit{Supplementary Materials for} \\ Rethinking the Role of Semantics in Large Language Models' In-Context Reasoning Abilities
}

\definecolor{codeblue}{rgb}{0.25,0.5,0.5}
\definecolor{codekw}{rgb}{0.85, 0.18, 0.50}
\definecolor{keywordgreen}{rgb}{0,0.6,0}
\lstset{
  backgroundcolor=\color{gray!10},
  basicstyle=\fontsize{8pt}{9pt}\ttfamily\selectfont,
  columns=fullflexible,
  breaklines=true,
  captionpos=b,
  commentstyle=\fontsize{7.5pt}{7.5pt}\color{codeblue},
  morekeywords = {system, user}, 
  keywordstyle = \bfseries,
}

%




\maketitle


 

\begin{appendices}

\DoToC




\section{Prompts}\label{prompts}
\subsection{Deductive reasoning}
 \subsubsection{Zero-Shot}

\begin{lstlisting}
system: You are a helpful assistant with deductive reasoning abilities. 
user: I will provide a set of logical rules L1 to L{number of rules} and facts F1 to F{number of basic facts}. Please select one single logical rule from L1 to L{number of rules} and a few facts from F1 to F{number of basic facts} to predict True/False of the unknown fact using deductive reasoning.
Logical rules: {logical rules}
Facts: {basic facts}
Unknown fact: {statement}
The answer (True or False) is:
\end{lstlisting}

\subsubsection{Zero-Shot-CoT}
\begin{lstlisting}
system: You are a helpful assistant with deductive reasoning abilities. Please select one single logical rule and a few facts to predict True/False of the following statement.
user: I will provide a set of logical rules L1 to L{number of rules} and facts F1 to F{number of basic facts}. Please select one single logical rule from L1 to L{number of rules} and a few facts from F1 to F{number of basic facts} to predict True/False of the following statement using deductive reasoning.
Logical rules: {logical rules}
Facts: {basic facts}
Statement: {statement}
Answer with True or False? Let's think step by step.

\end{lstlisting}

\subsubsection{Few-Shot-CoT}

\begin{lstlisting}
system: You are a helpful assistant with deductive reasoning abilities. Please select one single logical rule and a few facts to predict True/False of the following statement.
user:I will provide a set of logical rules L1 to L{number of rules} and facts F1 to F{number of basic facts}.
Logical rules: {logical rules}
Facts: {basic facts}
Please select one single logical rule from L1 to L{number of rules} and a few facts from F1 to F{number of basic facts} to predict True/False of the following statement using deductive reasoning.
{Demonstrations}
Statement: {statement}
Answer:  
\end{lstlisting}

\subsubsection{Examples of Few-Shot-CoT Demonstrations}

Statement: $r14$(Amelie, Jonathan)\\
Answer: We can use L11: $\forall A,B,C,D,E: r3(B, A) \land r3(B, C) \land r3(C, D) \land r3(D, E) \land r2(A) \rightarrow r14(A, E)$ to make a deduction. 
From the facts, we know that $r3$(Amelie, Thomas) and $r3$(Thomas, Jonathan). Therefore, we can apply L11 with A = Amelie, B = Thomas, C = Jonathan, D and E as variables. This gives us:
$r3$(Thomas, Jonathan) $\land$ $r3$(Amelie, Thomas) $\land$ $r3$(Thomas, D) $\land$ $r3$(D, E) $\land$ $r2$(Amelie) $\rightarrow$ $r14$(Amelie, E)
Since we know that $r2$(Amelie) is true from F7, we can simplify the above statement to:
$r3$(Thomas, Jonathan) $\land$ $r3$(Amelie, Thomas) $\land$ $r3$(Thomas, D) $\land$ $r3$(D, E) $\rightarrow$ $r14$(Amelie, E)
Now, we can substitute the facts $r3$(Amelie, Thomas) and $r3$(Thomas, Jonathan) to get:
$r3$(Thomas, Jonathan) $\land$ $r3$(Thomas, D) $\land$ $r3$(D, E) $\rightarrow$ $r14$(Amelie, E)
We can see that this statement is true if we choose D = Alina and E = Jonathan, since we know that $r3$(Thomas, Alina) from F50 and $r3$(Alina, Jonathan) from F56. Therefore, the statement $r14$(Amelie, Jonathan) is true. 
Answer: True.

Statement: $r31$(Philipp, Nina)\\
Answer: Let's use L28: $\forall A,B,C,D: r3(B, A) \land r3(C, B) \land r3(C, D) \land r1(A) \rightarrow r31(A, D)$ to see if we can prove the statement $r31$(Philipp, Nina).
From the facts, we know that $r3$(Luisa, Nina) (F52) and $r3$(Emilia, Philipp) (F62). We can use L28 with $A=Philipp$, $B=Emilia$, $C=Luisa$, and $D=Nina$ to get:
$r3$(Emilia, Philipp) $\land$ $r3$(Luisa, Emilia) $\land$ $r3$(Luisa, Nina) $\land$ $r1$(Philipp) $\rightarrow$ $r31$(Philipp, Nina)
Since all the conditions are met, we can conclude that $r31$(Philipp, Nina) is True. Therefore, the answer is True.

Statement: $r8$(Leonie, Nico)\\
Answer: We can use L5 to make a deduction about the statement $r8$(Leonie, Nico). L5 states that if there exist three individuals A, B, and C such that A is related to B, B is related to C, and A has a certain property (r2), then A is related to C in a certain way (r8). 
Using this rule, we can see that we have the following facts:
- $r3$(Leonie, Emily)\\
- $r3$(Emily, Nico)\\
Therefore, we can conclude that $r8$(Leonie, Nico) is true.

Statement: $r7$(Patrick, Alina)\\
Answer: We can use L4: $\forall A,B: r3(A, B) \land r1(A) \rightarrow r7(A, B)$ to determine if $r7$(Patrick, Alina) is true or false. 
From the facts, we know that $r3$(Patrick, Alina) is true (F55). We also know that $r1$(Patrick) is true (F9). Therefore, we can apply L4 to conclude that $r7$(Patrick, Alina) is true. 
Answer: True.

Statement: $r27$(Jonathan, Leonie)\\
Answer: Let's use L4: $\forall A,B: r3(A, B) \land r1(A) \rightarrow r7(A, B)$ and F56: $r3$(Patrick, Jonathan) to infer that $r7$(Patrick, Jonathan) is true. 
Then, we can use L24: $\forall A,B,C: r3(B, A) \land r3(C, B) \land r1(A) \rightarrow r27(A, C)$ and F44: $r3$(Leonie, Emily) to infer that $r27$(Jonathan, Leonie) is false, since there is no fact that supports $r3$(Jonathan, Emily). 
Therefore, the answer is False.


        
        

        
        

    

\subsubsection{Zero-Shot of removing rules setting}
\begin{lstlisting}
system: Please answer the question only with True or False. 
user: I will provide a set of facts. Please predict True/False of the unknown fact based on given facts.
Facts: {facts}
Unknown fact: {statement}
The answer (True or False) is: 
\end{lstlisting}

\subsection{Inductive reasoning}
    
\subsubsection{Zero-Shot}
\begin{lstlisting}
system: You are a helpful assistant with inductive reasoning abilities. Please generate one single rule to match the template and logically entail the facts. Note that the symbol '##' in the template should be filled with either 'r1' or 'r45', while the symbol '++' should be filled with either 'r43' or 'r44'.
user: I will give you a set of facts F1 to F{number of basic facts}, facts G1 to G{number of inferred fact} and a template for a logical rule. Please generate one single rule to match the template and logically entail the facts G1 to G{number of inferred fact} based on facts F1 to F{number of basic facts}.
Facts: {facts} 
Template: {rule template} 
Note that the symbol '##' in the template should be filled with either 'r1' or 'r45', while the symbol '++' should be filled with either 'r43' or 'r44'.
After filling in the template, the generated rule is:

\end{lstlisting}
    
\subsubsection{Zero-Shot CoT}
\begin{lstlisting}
system: You are a helpful assistant with inductive reasoning abilities. Please generate one single rule to match the template and logically entail the facts. Note that the symbol '##' in the template should be filled with either 'r1' or 'r45', while the symbol '++' should be filled with either 'r43' or 'r44'.
user: I will give you a set of facts F1 to F{number of basic facts}, facts G1 to G{number of inferred fact} and a template for a logical rule. Please generate one single rule to match the template and logically entail the facts G1 to G{number of inferred fact} based on facts F1 to F{number of basic facts}.
Facts: {facts} 
Template: {rule template} 
Note that the symbol '##' in the template should be filled with either 'r1' or 'r45', while the symbol '++' should be filled with either 'r43' or 'r44'.
After filling in the template, the generated rule is: Let's think step by step.

\end{lstlisting}
    
\subsubsection{Zero-Shot of removing facts setting}
\begin{lstlisting}
system: Please generate one single rule to match the template. Note that the symbol '##' in the template should be filled with either 'parent' or 'child', while the symbol '++' should be filled with either 'male' or 'female'.
user: I will give you a template for a logical rule. Please generate one single rule to match the template and logically infer the relation sister
Template: If A is ## of B and B is ## of C and A is ++, then A is sister of C.
Note that the symbol '##' in the template should be filled with either 'parent' or 'child', while the symbol '++' should be filled with either 'male' or 'female'.
After filling in the template, the generated rule is: 
\end{lstlisting}

\subsection{Abductive reasoning}
    
\subsubsection{Zero-Shot}
\begin{lstlisting}
system: You are a helpful assistant with abductive reasoning abilities. Please select one single logical rule and a few facts to explain the following statement.
user: I will provide a set of logical rules L1 to L{number of rules} and facts F1 to F{number of basic facts}. Please select one single logical rule from L1 to L{number of rules} and a few facts from F1 to F{number of basic facts} to explain the following statement.
Rules: {logical rules}
Facts: {basic facts}
Statement: {statement}
Answer with the numbers of the selected rule and facts. The selected rule and facts are:
\end{lstlisting}

\subsubsection{Zero-Shot-CoT}
\begin{lstlisting}
system: You are a helpful assistant with abductive reasoning abilities. Please select one single logical rule and a few facts to explain the following statement.
user: I will provide a set of logical rules L1 to L{number of rules} and facts F1 to F{number of basic facts}. Please select one single logical rule from L1 to L{number of rules} and a few facts from F1 to F{number of basic facts} to explain the following statement.
Rules: {logical rules}
Facts: {basic facts}
Statement: {statement}
Answer with the numbers of the selected rule and facts. The selected rule and facts are: Let's think step by step.
\end{lstlisting}
\subsubsection{Few-Shot-CoT prompt}

\begin{lstlisting}
system: You are a helpful assistant with abductive reasoning abilities. Please select one single logical rule and a few facts to explain the following statement.
user: 
    I will provide a set of logical rules L1 to L{number of rules} and facts F1 to F{number of basic facts}. Please select one single logical rule from L1 to L{number of rules} and a few facts from F1 to F{number of basic facts} to explain the following statement.
    Rules: {logical rules}
    Facts: {basic facts}
    {Demonstrations} 
    Statement: {statement}
    Answer:

\end{lstlisting}

\subsubsection{Examples of Few-Shot-CoT Demonstrations}
    
Statement: $r4$(Laura, Fabian)\\
Answer: To explain the statement $r4$(Laura, Fabian), we can use the rule L3 and fact F1 and F37. \\
L3: $\forall A,B: r1(A, B) \land r44(A) \rightarrow r4(A, B)$ \\
F1: $r1$(Laura,Fabian) \\
F37: $r44$(Laura) \\
According to L3, if there is a relationship r1 between A and B, and A has the property r44, then the relationship r4 holds between A and B. In this case, we have a relationship r1 between Laura and Fabian (F1) and Laura has the property r44 (F37). Therefore, the relationship r4 holds between Laura and Fabian, which is the given statement. \\
The selected rule and facts are: L3, F1, and F37.

Statement: $r5$(Stefan, Emily)\\
Answer: To explain the statement $r5$(Stefan, Emily), we can use the rule L4 and fact F17.
L4: $\forall A,B: r1(A, B) \land r43(A) \rightarrow r5(A, B)$\\
F17: $r1$(Stefan,Emily)\\
According to rule L4, if there is a relation r1(A, B) and A has the property r43, then the relation r5(A, B) holds. In our case, we have the fact F17, which states that there is a relation r1(Stefan, Emily). Additionally, Stefan has the property r43 (F61). Therefore, we can conclude that the relation r5(Stefan, Emily) holds.\\
The selected rule and facts are: L4, F17, and F61.

Statement: $r23$(Felix, Elias)\\
Answer: To explain the statement $r23$(Felix, Elias), we can use the rule L22 and fact F5.
L22: $\forall A,B: r1(B, A) \land r43(A) \rightarrow r23(A, B)$\\
F5: $r1$(Elias,Felix)\\
According to rule L22, if there exists a relationship $r1$(B, A) and A has the property $r43$, then the relationship $r23$(A, B) holds. In our case, we have the fact F5, which states that there is a relationship $r1$(Elias, Felix). Additionally, we know that Felix has the property $r43$ (F59). Therefore, applying rule L22, we can conclude that the relationship $r23$(Felix, Elias) holds.\\
The selected rule and facts are: L22, F5, and F59.

Statement: $r3$(Felix, Fabian)\\
Answer: To explain the statement $r3$(Felix, Fabian), we can use the rule L2 and facts F2, F1, and F59.\\
L2: $\forall A,B,C: r1(B, A) \land r1(B, C) \land r43(A) \rightarrow r3(A, C)$\\
F1: $r1$(Laura,Fabian)\\
F2: $r1$(Laura,Felix)\\
F59: $r43$(Felix)\\
By substituting the values into L2, we get:\\
$r1$(Laura, Felix) $\land$ $r1$(Laura, Fabian) $\land$ $r43$(Felix) $\rightarrow$ $r3$(Felix, Fabian)\\
Since all the premises are true according to the given facts, the conclusion $r3$(Felix, Fabian) is also true.\\
The selected rule and facts are: L2, F1, F2, and F59.

Statement: $r7$(Fabian, Helga)\\
Answer: To explain the statement $r7$(Fabian, Helga), we can use the rule L6 and facts F13, F21, and F39.\\
L6: $\forall A,B,C: r1(A, B) \land r1(B, C) \land r43(A) \rightarrow r7(A, C)$\\
F13: $r1$(Fabian,Thomas)\\
F21: $r1$(Thomas,Helga)\\
F39: $r43$(Fabian)\\
According to L6, if there exist A, B, and C such that $r1(A, B)$, $r1(B, C)$, and $r43(A)$, then $r7(A, C)$ holds. In our case, A = Fabian, B = Thomas, and C = Helga. Since we have $r1$(Fabian,Thomas) from F13, $r1$(Thomas,Helga) from F21, and $r43$(Fabian) from F39, we can conclude that $r7$(Fabian, Helga) is true.\\
So, the selected rule and facts are: L6, F13, F21, and F39.

\subsubsection{Few-Shot CoT of removing rules setting} 
\begin{lstlisting}
system: You are a helpful assistant with abductive reasoning abilities. Please select a few facts to explain the following statement. 
user: I will provide a set of facts F1 to F63. Please select a few facts from F1 to F63 to explain the following statement. 
Facts: {facts}

Statement: Laura is mother of Felix.
Answer: To explain the statement "Laura is mother of Felix", we can use Facts:
Fact F2 states: Laura is parent of Felix.
Fact F37 states: Laura is female.
Using F2 and F37, we can conclude that "Laura is mother of Felix" holds.
Therefore, the selected rule and facts are F2, F37.

Statement: Samuel is brother of Alina.
Answer: To infer the statement "Samuel is brother of Alina", we have:
F27: Patrick is parent of Samuel.
F28: Patrick is parent of Alina.
F47: Samuel is male.
Based on these facts, we can infer "Samuel is brother of Alina":
Therefore, the selected rule and facts are F27, F28, F47.

Statement: Patrick is grandfather of David.
Answer: To explain the statement "Patrick is grandfather of David", we have:
F28: Patrick is parent of Alina.
F7: Alina is parent of David.
F45: Patrick is male.
Based on these facts, we can infer "Patrick is grandfather of David":
Therefore, the selected rule and facts are F28, F7, F45.

Statement: Amelie is daughter of Elena.
Answer: To explain the statement "Amelie is daughter of Elena", we have:
F20: Elena is parent of Amelie.
F43: Amelie is female.
Based on these facts, we can infer "Amelie is daughter of Elena".
Therefore, the selected rule and facts are F20, F43.

Statement: Claudia is sister of Felix
Answer: To prove the statement "Claudia is sister of Felix", we can use facts:
F3: Laura is parent of Claudia.
F2: Laura is parent of Felix.
F40: Claudia is female.
Based on these facts, we can infer "Claudia is sister of Felix".
Therefore, the selected rule and facts are F3, F2, F40.

Statement: Laura is mother of Fabian.
Answer: 
\end{lstlisting}




            
            




            
            

\subsection{Commonsense reasoning}

\subsubsection{Zero-Shot prompt}
\begin{lstlisting}
system: Please choose answer from [A, B, C, D, E] and output one capital letter only.
user: Question: {Question}
{Multiple choices}
Answer:
\end{lstlisting}

\subsubsection{Few-Shot prompt}
\begin{lstlisting}
system: Please choose answer from [A, B, C, D, E] and output one capital letter only.
user: {Demonstrations:}
Question: {Question}
{Multiple choices}
Answer:
\end{lstlisting}

\subsubsection{Examples of Few-Shot Demonstrations}

\textbf{Standard: } \\
Question: When is hotter?\\
Choices: A.desert B.night C.morning D.noon \\
Answer: A.

\textbf{Predict-explain:}\\
Question: When is hotter?\\
Choices: A.desert B.night C.morning D.noon\\
Answer: A. The desert is hotter than the other options listed because it receives direct sunlight and has little to no vegetation to provide shade or cool the air. During the day, the sun beats down on the desert, causing temperatures to rise significantly. At night, the lack of vegetation and moisture in the air means that heat is quickly lost, causing temperatures to drop rapidly. In the morning and at noon, the sun is still high in the sky, but temperatures may be slightly cooler due to the presence of some shade and cooler air from the night before.

\textbf{Expain-predict: }\\
Question: When is hotter?\\
Choices: A.desert B.night C.morning D.noon \\
The desert is hotter than the other options listed because it receives direct sunlight and has little to no vegetation to provide shade or cool the air. During the day, the sun beats down on the desert, causing temperatures to rise significantly. At night, the lack of vegetation and moisture in the air means that heat is quickly lost, causing temperatures to drop rapidly. In the morning and at noon, the sun is still high in the sky, but temperatures may be slightly cooler due to the presence of some shade and cooler air from the night before. Answer: A.

\subsection{LLaMA Fine-tuning Prompt}\label{finetune_prompt}
\begin{lstlisting}
Below is an instruction that describes a task, paired with an input that provides further context.
Write a response that appropriately completes the request.
Instruction: {Head} is the {Relation} of {Tail}
Input: {input} 
Response: 
\end{lstlisting}


\section{Deduction examples of Symbolic Tree datasets }
In this section, we provide examples of deduction experiments conducted on the Symbolic Tree datasets. We present examples for both the \textit{Semantics} and \textit{Symbols} settings, represented in both natural language text and logic language

\subsection{Semantics}
\subsubsection{Logic language representations}
\begin{lstlisting}
Logical rules:
L1: $\forall A,B,C: parentOf(B, A) \land parentOf(B, C) \land female(A) \rightarrow sisterOf(A,C)$
L2: $\forall A,B,C: parentOf(B, A) \land parentOf(B, C) \land male(A) \rightarrow brotherOf(A,C)$
L3: $\forall A,B: parentOf(A, B) \land female(A) \rightarrow motherOf(A,B)$
L4: $\forall A,B: parentOf(A, B) \land male(A) \rightarrow fatherOf(A,B)$
L5: $\forall A,B,C: parentOf(A, B) \land parentOf(B, C) \land female(A) \rightarrow grandmotherOf(A,C)$
L6: $\forall A,B,C: parentOf(A, B) \land parentOf(B, C) \land male(A) \rightarrow grandfatherOf(A,C)$
L7: $\forall A,B,C,D: parentOf(A, B) \land parentOf(B, C) \land parentOf(C, D) \land female(A) \rightarrow greatGrandmotherOf(A,D)$
L8: $\forall A,B,C,D: parentOf(A, B) \land parentOf(B, C) \land parentOf(C, D) \land male(A) \rightarrow greatGrandfatherOf(A,D)$
L9: $\forall A,B,C,D: parentOf(B, A) \land parentOf(B, C) \land parentOf(C, D) \land female(A) \rightarrow auntOf(A,D)$
L10: $\forall A,B,C,D: parentOf(B, A) \land parentOf(B, C) \land parentOf(C, D) \land male(A) \rightarrow uncleOf(A,D)$
L11: $\forall A,B,C,D,E: parentOf(B, A) \land parentOf(B, C) \land parentOf(C, D) \land parentOf(D, E) \land female(A) \rightarrow greatAuntOf(A,E)$
L12: $\forall A,B,C,D,E: parentOf(B, A) \land parentOf(B, C) \land parentOf(C, D) \land parentOf(D, E) \land male(A) \rightarrow greatUncleOf(A,E)$
L13: $\forall A,B,C,D,E,F: parentOf(B, A) \land parentOf(C, B) \land parentOf(C, D) \land parentOf(D, E) \land parentOf(E, F) \land female(A) \rightarrow secondAuntOf(A,F)$
L14: $\forall A,B,C,D,E,F: parentOf(B, A) \land parentOf(C, B) \land parentOf(C, D) \land parentOf(D, E) \land parentOf(E, F) \land male(A) \rightarrow secondUncleOf(A,F)$
L15: $\forall A,B,C,D,E: parentOf(B, A) \land parentOf(C, B) \land parentOf(C, D) \land parentOf(D, E) \land female(A) \rightarrow girlCousinOf(A,E)$
L16: $\forall A,B,C,D,E: parentOf(B, A) \land parentOf(C, B) \land parentOf(C, D) \land parentOf(D, E) \land male(A) \rightarrow boyCousinOf(A,E)$
L17: $\forall A,B,C,D,E,F,G: parentOf(B, A) \land parentOf(C, B) \land parentOf(D, C) \land parentOf(D, E) \land parentOf(E, F) \land parentOf(F, G) \land female(A) \rightarrow girlSecondCousinOf(A,G)$
L18: $\forall A,B,C,D,E,F,G: parentOf(B, A) \land parentOf(C, B) \land parentOf(D, C) \land parentOf(D, E) \land parentOf(E, F) \land parentOf(F, G) \land male(A) \rightarrow boySecondCousinOf(A,G)$
L19: $\forall A,B,C,D,E,F: parentOf(B, A) \land parentOf(C, B) \land parentOf(D, C) \land parentOf(D, E) \land parentOf(E, F) \land female(A) \rightarrow girlFirstCousinOnceRemovedOf(A,F)$
L20: $\forall A,B,C,D,E,F: parentOf(B, A) \land parentOf(C, B) \land parentOf(D, C) \land parentOf(D, E) \land parentOf(E, F) \land male(A) \rightarrow boyFirstCousinOnceRemovedOf(A,F)$
L21: $\forall A,B: parentOf(B, A) \land female(A) \rightarrow daughterOf(A,B)$
L22: $\forall A,B: parentOf(B, A) \land male(A) \rightarrow sonOf(A,B)$
L23: $\forall A,B,C: parentOf(B, A) \land parentOf(C, B) \land female(A) \rightarrow granddaughterOf(A,C)$
L24: $\forall A,B,C: parentOf(B, A) \land parentOf(C, B) \land male(A) \rightarrow grandsonOf(A,C)$
L25: $\forall A,B,C,D: parentOf(B, A) \land parentOf(C, B) \land parentOf(D, C) \land female(A) \rightarrow greatGranddaughterOf(A,D)$
L26: $\forall A,B,C,D: parentOf(B, A) \land parentOf(C, B) \land parentOf(D, C) \land male(A) \rightarrow greatGrandsonOf(A,D)$
L27: $\forall A,B,C,D: parentOf(B, A) \land parentOf(C, B) \land parentOf(C, D) \land female(A) \rightarrow nieceOf(A,D)$
L28: $\forall A,B,C,D: parentOf(B, A) \land parentOf(C, B) \land parentOf(C, D) \land male(A) \rightarrow nephewOf(A,D)$

Facts:
F1: female(Laura)
F2: male(Elias)
F3: male(Fabian)
F4: female(Claudia)
F5: female(Elena)
F6: male(Thomas)
F7: female(Amelie)
F8: female(Luisa)
F9: male(Patrick)
F10: female(Emilia)
F11: male(Samuel)
F12: female(Alina)
F13: male(Jonathan)
F14: male(Philipp)
F15: male(Nico)
F16: male(David)
F17: female(Emily)
F18: male(Konstantin)
F19: male(Florian)
F20: female(Helga)
F21: female(Nina)
F22: female(Lea)
F23: male(Felix)
F24: female(Leonie)
F25: male(Stefan)
F26: male(Gabriel)
F27: male(Tobias)
F28: parentOf(Laura, Fabian)
F29: parentOf(Laura, Felix)
F30: parentOf(Laura, Claudia)
F31: parentOf(Elias, Fabian)
F32: parentOf(Elias, Felix)
F33: parentOf(Elias, Claudia)
F34: parentOf(Alina, David)
F35: parentOf(Alina, Lea)
F36: parentOf(Nico, David)
F37: parentOf(Nico, Lea)
F38: parentOf(Emily, Nico)
F39: parentOf(Konstantin, Nico)
F40: parentOf(Fabian, Thomas)
F41: parentOf(Fabian, Amelie)
F42: parentOf(Nina, Tobias)
F43: parentOf(Leonie, Emily)
F44: parentOf(Stefan, Emily)
F45: parentOf(Gabriel, Tobias)
F46: parentOf(Elena, Thomas)
F47: parentOf(Elena, Amelie)
F48: parentOf(Thomas, Helga)
F49: parentOf(Thomas, Nina)
F50: parentOf(Thomas, Patrick)
F51: parentOf(Luisa, Helga)
F52: parentOf(Luisa, Nina)
F53: parentOf(Luisa, Patrick)
F54: parentOf(Patrick, Samuel)
F55: parentOf(Patrick, Alina)
F56: parentOf(Patrick, Jonathan)
F57: parentOf(Patrick, Philipp)
F58: parentOf(Patrick, Florian)
F59: parentOf(Emilia, Samuel)
F60: parentOf(Emilia, Alina)
F61: parentOf(Emilia, Jonathan)
F62: parentOf(Emilia, Philipp)
F63: parentOf(Emilia, Florian)

Unknown fact: boyCousinOf(Tobias, David)
\end{lstlisting}

\subsubsection{Natural language representations}
\begin{lstlisting}
Logical rules:
L1: If B is parent of A and B is parent of C and A is female, then A is sister of D.
L2: If B is parent of A and B is parent of C and A is male, then A is brother of D.
L3: If A is parent of B and A is female, then A is mother of C.
L4: If A is parent of B and A is male, then A is father of C.
L5: If A is parent of B and B is parent of C and A is female, then A is grandmother of D.
L6: If A is parent of B and B is parent of C and A is male, then A is grandfather of D.
L7: If A is parent of B and B is parent of C and C is parent of D and A is female, then A is greatGrandmother of E.
L8: If A is parent of B and B is parent of C and C is parent of D and A is male, then A is greatGrandfather of E.
L9: If B is parent of A and B is parent of C and C is parent of D and A is female, then A is aunt of E.
L10: If B is parent of A and B is parent of C and C is parent of D and A is male, then A is uncle of E.
L11: If B is parent of A and B is parent of C and C is parent of D and D is parent of E and A is female, then A is greatAunt of F.
L12: If B is parent of A and B is parent of C and C is parent of D and D is parent of E and A is male, then A is greatUncle of F.
L13: If B is parent of A and C is parent of B and C is parent of D and D is parent of E and E is parent of F and A is female, then A is secondAunt of G.
L14: If B is parent of A and C is parent of B and C is parent of D and D is parent of E and E is parent of F and A is male, then A is secondUncle of G.
L15: If B is parent of A and C is parent of B and C is parent of D and D is parent of E and A is female, then A is girlCousin of F.
L16: If B is parent of A and C is parent of B and C is parent of D and D is parent of E and A is male, then A is boyCousin of F.
L17: If B is parent of A and C is parent of B and D is parent of C and D is parent of E and E is parent of F and F is parent of G and A is female, then A is girlSecondCousin of H.
L18: If B is parent of A and C is parent of B and D is parent of C and D is parent of E and E is parent of F and F is parent of G and A is male, then A is boySecondCousin of H.
L19: If B is parent of A and C is parent of B and D is parent of C and D is parent of E and E is parent of F and A is female, then A is girlFirstCousinOnceRemoved of G.
L20: If B is parent of A and C is parent of B and D is parent of C and D is parent of E and E is parent of F and A is male, then A is boyFirstCousinOnceRemoved of G.
L21: If B is parent of A and A is female, then A is daughter of C.
L22: If B is parent of A and A is male, then A is son of C.
L23: If B is parent of A and C is parent of B and A is female, then A is granddaughter of D.
L24: If B is parent of A and C is parent of B and A is male, then A is grandson of D.
L25: If B is parent of A and C is parent of B and D is parent of C and A is female, then A is greatGranddaughter of E.
L26: If B is parent of A and C is parent of B and D is parent of C and A is male, then A is greatGrandson of E.
L27: If B is parent of A and C is parent of B and C is parent of D and A is female, then A is niece of E.
L28: If B is parent of A and C is parent of B and C is parent of D and A is male, then A is nephew of E.

Facts:
F1: Laura is female.
F2: Elias is male.
F3: Fabian is male.
F4: Claudia is female.
F5: Elena is female.
F6: Thomas is male.
F7: Amelie is female.
F8: Luisa is female.
F9: Patrick is male.
F10: Emilia is female.
F11: Samuel is male.
F12: Alina is female.
F13: Jonathan is male.
F14: Philipp is male.
F15: Nico is male.
F16: David is male.
F17: Emily is female.
F18: Konstantin is male.
F19: Florian is male.
F20: Helga is female.
F21: Nina is female.
F22: Lea is female.
F23: Felix is male.
F24: Leonie is female.
F25: Stefan is male.
F26: Gabriel is male.
F27: Tobias is male.
F28: Laura is parent of Fabian.
F29: Laura is parent of Felix.
F30: Laura is parent of Claudia.
F31: Elias is parent of Fabian.
F32: Elias is parent of Felix.
F33: Elias is parent of Claudia.
F34: Alina is parent of David.
F35: Alina is parent of Lea.
F36: Nico is parent of David.
F37: Nico is parent of Lea.
F38: Emily is parent of Nico.
F39: Konstantin is parent of Nico.
F40: Fabian is parent of Thomas.
F41: Fabian is parent of Amelie.
F42: Nina is parent of Tobias.
F43: Leonie is parent of Emily.
F44: Stefan is parent of Emily.
F45: Gabriel is parent of Tobias.
F46: Elena is parent of Thomas.
F47: Elena is parent of Amelie.
F48: Thomas is parent of Helga.
F49: Thomas is parent of Nina.
F50: Thomas is parent of Patrick.
F51: Luisa is parent of Helga.
F52: Luisa is parent of Nina.
F53: Luisa is parent of Patrick.
F54: Patrick is parent of Samuel.
F55: Patrick is parent of Alina.
F56: Patrick is parent of Jonathan.
F57: Patrick is parent of Philipp.
F58: Patrick is parent of Florian.
F59: Emilia is parent of Samuel.
F60: Emilia is parent of Alina.
F61: Emilia is parent of Jonathan.
F62: Emilia is parent of Philipp.
F63: Emilia is parent of Florian.

Unknown fact: Gabriel is uncle of Lea.
\end{lstlisting}

\subsection{Symbolization}
\subsubsection{Logic language representations}
\begin{lstlisting}
Logical rules:
L1: $\forall A,B,C: r3(B, A) \land r3(B, C) \land r2(A) \rightarrow r4(A, C)$
L2: $\forall A,B,C: r3(B, A) \land r3(B, C) \land r1(A) \rightarrow r5(A, C)$
L3: $\forall A,B: r3(A, B) \land r2(A) \rightarrow r6(A, B)$
L4: $\forall A,B: r3(A, B) \land r1(A) \rightarrow r7(A, B)$
L5: $\forall A,B,C: r3(A, B) \land r3(B, C) \land r2(A) \rightarrow r8(A, C)$
L6: $\forall A,B,C: r3(A, B) \land r3(B, C) \land r1(A) \rightarrow r9(A, C)$
L7: $\forall A,B,C,D: r3(A, B) \land r3(B, C) \land r3(C, D) \land r2(A) \rightarrow r10(A, D)$
L8: $\forall A,B,C,D: r3(A, B) \land r3(B, C) \land r3(C, D) \land r1(A) \rightarrow r11(A, D)$
L9: $\forall A,B,C,D: r3(B, A) \land r3(B, C) \land r3(C, D) \land r2(A) \rightarrow r12(A, D)$
L10: $\forall A,B,C,D: r3(B, A) \land r3(B, C) \land r3(C, D) \land r1(A) \rightarrow r13(A, D)$
L11: $\forall A,B,C,D,E: r3(B, A) \land r3(B, C) \land r3(C, D) \land r3(D, E) \land r2(A) \rightarrow r14(A, E)$
L12: $\forall A,B,C,D,E: r3(B, A) \land r3(B, C) \land r3(C, D) \land r3(D, E) \land r1(A) \rightarrow r15(A, E)$
L13: $\forall A,B,C,D,E,F: r3(B, A) \land r3(C, B) \land r3(C, D) \land r3(D, E) \land r3(E, F) \land r2(A) \rightarrow r16(A, F)$
L14: $\forall A,B,C,D,E,F: r3(B, A) \land r3(C, B) \land r3(C, D) \land r3(D, E) \land r3(E, F) \land r1(A) \rightarrow r17(A, F)$
L15: $\forall A,B,C,D,E: r3(B, A) \land r3(C, B) \land r3(C, D) \land r3(D, E) \land r2(A) \rightarrow r18(A, E)$
L16: $\forall A,B,C,D,E: r3(B, A) \land r3(C, B) \land r3(C, D) \land r3(D, E) \land r1(A) \rightarrow r19(A, E)$
L17: $\forall A,B,C,D,E,F,G: r3(B, A) \land r3(C, B) \land r3(D, C) \land r3(D, E) \land r3(E, F) \land r3(F, G) \land r2(A) \rightarrow r20(A, G)$
L18: $\forall A,B,C,D,E,F,G: r3(B, A) \land r3(C, B) \land r3(D, C) \land r3(D, E) \land r3(E, F) \land r3(F, G) \land r1(A) \rightarrow r21(A, G)$
L19: $\forall A,B,C,D,E,F: r3(B, A) \land r3(C, B) \land r3(D, C) \land r3(D, E) \land r3(E, F) \land r2(A) \rightarrow r22(A, F)$
L20: $\forall A,B,C,D,E,F: r3(B, A) \land r3(C, B) \land r3(D, C) \land r3(D, E) \land r3(E, F) \land r1(A) \rightarrow r23(A, F)$
L21: $\forall A,B: r3(B, A) \land r2(A) \rightarrow r24(A, B)$
L22: $\forall A,B: r3(B, A) \land r1(A) \rightarrow r25(A, B)$
L23: $\forall A,B,C: r3(B, A) \land r3(C, B) \land r2(A) \rightarrow r26(A, C)$
L24: $\forall A,B,C: r3(B, A) \land r3(C, B) \land r1(A) \rightarrow r27(A, C)$
L25: $\forall A,B,C,D: r3(B, A) \land r3(C, B) \land r3(D, C) \land r2(A) \rightarrow r28(A, D)$
L26: $\forall A,B,C,D: r3(B, A) \land r3(C, B) \land r3(D, C) \land r1(A) \rightarrow r29(A, D)$
L27: $\forall A,B,C,D: r3(B, A) \land r3(C, B) \land r3(C, D) \land r2(A) \rightarrow r30(A, D)$
L28: $\forall A,B,C,D: r3(B, A) \land r3(C, B) \land r3(C, D) \land r1(A) \rightarrow r31(A, D)$

Facts:
F1: $r2$(Laura)
F2: $r1$(Elias)
F3: $r1$(Fabian)
F4: $r2$(Claudia)
F5: $r2$(Elena)
F6: $r1$(Thomas)
F7: $r2$(Amelie)
F8: $r2$(Luisa)
F9: $r1$(Patrick)
F10: $r2$(Emilia)
F11: $r1$(Samuel)
F12: $r2$(Alina)
F13: $r1$(Jonathan)
F14: $r1$(Philipp)
F15: $r1$(Nico)
F16: $r1$(David)
F17: $r2$(Emily)
F18: $r1$(Konstantin)
F19: $r1$(Florian)
F20: $r2$(Helga)
F21: $r2$(Nina)
F22: $r2$(Lea)
F23: $r1$(Felix)
F24: $r2$(Leonie)
F25: $r1$(Stefan)
F26: $r1$(Gabriel)
F27: $r1$(Tobias)
F28: $r3$(Laura, Fabian)
F29: $r3$(Laura, Felix)
F30: $r3$(Laura, Claudia)
F31: $r3$(Elias, Fabian)
F32: $r3$(Elias, Felix)
F33: $r3$(Elias, Claudia)
F34: $r3$(Alina, David)
F35: $r3$(Alina, Lea)
F36: $r3$(Nico, David)
F37: $r3$(Nico, Lea)
F38: $r3$(Emily, Nico)
F39: $r3$(Konstantin, Nico)
F40: $r3$(Fabian, Thomas)
F41: $r3$(Fabian, Amelie)
F42: $r3$(Nina, Tobias)
F43: $r3$(Leonie, Emily)
F44: $r3$(Stefan, Emily)
F45: $r3$(Gabriel, Tobias)
F46: $r3$(Elena, Thomas)
F47: $r3$(Elena, Amelie)
F48: $r3$(Thomas, Helga)
F49: $r3$(Thomas, Nina)
F50: $r3$(Thomas, Patrick)
F51: $r3$(Luisa, Helga)
F52: $r3$(Luisa, Nina)
F53: $r3$(Luisa, Patrick)
F54: $r3$(Patrick, Samuel)
F55: $r3$(Patrick, Alina)
F56: $r3$(Patrick, Jonathan)
F57: $r3$(Patrick, Philipp)
F58: $r3$(Patrick, Florian)
F59: $r3$(Emilia, Samuel)
F60: $r3$(Emilia, Alina)
F61: $r3$(Emilia, Jonathan)
F62: $r3$(Emilia, Philipp)
F63: $r3$(Emilia, Florian)

Unknown fact: $r9$(Thomas, Claudia)
\end{lstlisting}

\subsubsection{Natural language representations:}
\begin{lstlisting}
Logical rules:
L1: If B is $r3$ of A and B is $r3$ of C and A is $r2$, then A is $r4$ of D.
L2: If B is $r3$ of A and B is $r3$ of C and A is $r1$, then A is $r5$ of D.
L3: If A is $r3$ of B and A is $r2$, then A is $r6$ of C.
L4: If A is $r3$ of B and A is $r1$, then A is $r7$ of C.
L5: If A is $r3$ of B and B is $r3$ of C and A is $r2$, then A is $r8$ of D.
L6: If A is $r3$ of B and B is $r3$ of C and A is $r1$, then A is $r9$ of D.
L7: If A is $r3$ of B and B is $r3$ of C and C is $r3$ of D and A is $r2$, then A is $r10$ of E.
L8: If A is $r3$ of B and B is $r3$ of C and C is $r3$ of D and A is $r1$, then A is $r11$ of E.
L9: If B is $r3$ of A and B is $r3$ of C and C is $r3$ of D and A is $r2$, then A is $r12$ of E.
L10: If B is $r3$ of A and B is $r3$ of C and C is $r3$ of D and A is $r1$, then A is $r13$ of E.
L11: If B is $r3$ of A and B is $r3$ of C and C is $r3$ of D and D is $r3$ of E and A is $r2$, then A is $r14$ of F.
L12: If B is $r3$ of A and B is $r3$ of C and C is $r3$ of D and D is $r3$ of E and A is $r1$, then A is $r15$ of F.
L13: If B is $r3$ of A and C is $r3$ of B and C is $r3$ of D and D is $r3$ of E and E is $r3$ of F and A is $r2$, then A is $r16$ of G.
L14: If B is $r3$ of A and C is $r3$ of B and C is $r3$ of D and D is $r3$ of E and E is $r3$ of F and A is $r1$, then A is $r17$ of G.
L15: If B is $r3$ of A and C is $r3$ of B and C is $r3$ of D and D is $r3$ of E and A is $r2$, then A is $r18$ of F.
L16: If B is $r3$ of A and C is $r3$ of B and C is $r3$ of D and D is $r3$ of E and A is $r1$, then A is $r19$ of F.
L17: If B is $r3$ of A and C is $r3$ of B and D is $r3$ of C and D is $r3$ of E and E is $r3$ of F and F is $r3$ of G and A is $r2$, then A is $r20$ of H.
L18: If B is $r3$ of A and C is $r3$ of B and D is $r3$ of C and D is $r3$ of E and E is $r3$ of F and F is $r3$ of G and A is $r1$, then A is $r21$ of H.
L19: If B is $r3$ of A and C is $r3$ of B and D is $r3$ of C and D is $r3$ of E and E is $r3$ of F and A is $r2$, then A is $r22$ of G.
L20: If B is $r3$ of A and C is $r3$ of B and D is $r3$ of C and D is $r3$ of E and E is $r3$ of F and A is $r1$, then A is $r23$ of G.
L21: If B is $r3$ of A and A is $r2$, then A is $r24$ of C.
L22: If B is $r3$ of A and A is $r1$, then A is $r25$ of C.
L23: If B is $r3$ of A and C is $r3$ of B and A is $r2$, then A is $r26$ of D.
L24: If B is $r3$ of A and C is $r3$ of B and A is $r1$, then A is $r27$ of D.
L25: If B is $r3$ of A and C is $r3$ of B and D is $r3$ of C and A is $r2$, then A is $r28$ of E.
L26: If B is $r3$ of A and C is $r3$ of B and D is $r3$ of C and A is $r1$, then A is $r29$ of E.
L27: If B is $r3$ of A and C is $r3$ of B and C is $r3$ of D and A is $r2$, then A is $r30$ of E.
L28: If B is $r3$ of A and C is $r3$ of B and C is $r3$ of D and A is $r1$, then A is $r31$ of E.

Facts:
F1: Laura is $r2$.
F2: Elias is $r1$.
F3: Fabian is $r1$.
F4: Claudia is $r2$.
F5: Elena is $r2$.
F6: Thomas is $r1$.
F7: Amelie is $r2$.
F8: Luisa is $r2$.
F9: Patrick is $r1$.
F10: Emilia is $r2$.
F11: Samuel is $r1$.
F12: Alina is $r2$.
F13: Jonathan is $r1$.
F14: Philipp is $r1$.
F15: Nico is $r1$.
F16: David is $r1$.
F17: Emily is $r2$.
F18: Konstantin is $r1$.
F19: Florian is $r1$.
F20: Helga is $r2$.
F21: Nina is $r2$.
F22: Lea is $r2$.
F23: Felix is $r1$.
F24: Leonie is $r2$.
F25: Stefan is $r1$.
F26: Gabriel is $r1$.
F27: Tobias is $r1$.
F28: Laura is $r3$ of Fabian.
F29: Laura is $r3$ of Felix.
F30: Laura is $r3$ of Claudia.
F31: Elias is $r3$ of Fabian.
F32: Elias is $r3$ of Felix.
F33: Elias is $r3$ of Claudia.
F34: Alina is $r3$ of David.
F35: Alina is $r3$ of Lea.
F36: Nico is $r3$ of David.
F37: Nico is $r3$ of Lea.
F38: Emily is $r3$ of Nico.
F39: Konstantin is $r3$ of Nico.
F40: Fabian is $r3$ of Thomas.
F41: Fabian is $r3$ of Amelie.
F42: Nina is $r3$ of Tobias.
F43: Leonie is $r3$ of Emily.
F44: Stefan is $r3$ of Emily.
F45: Gabriel is $r3$ of Tobias.
F46: Elena is $r3$ of Thomas.
F47: Elena is $r3$ of Amelie.
F48: Thomas is $r3$ of Helga.
F49: Thomas is $r3$ of Nina.
F50: Thomas is $r3$ of Patrick.
F51: Luisa is $r3$ of Helga.
F52: Luisa is $r3$ of Nina.
F53: Luisa is $r3$ of Patrick.
F54: Patrick is $r3$ of Samuel.
F55: Patrick is $r3$ of Alina.
F56: Patrick is $r3$ of Jonathan.
F57: Patrick is $r3$ of Philipp.
F58: Patrick is $r3$ of Florian.
F59: Emilia is $r3$ of Samuel.
F60: Emilia is $r3$ of Alina.
F61: Emilia is $r3$ of Jonathan.
F62: Emilia is $r3$ of Philipp.
F63: Emilia is $r3$ of Florian.

Unknown fact: Nico is $r27$ of Stefan.
\end{lstlisting}

\subsection{Semantics of removing rule setting}
\begin{lstlisting}
I will provide a set of facts. Please predict True/False of the unknown fact based on given facts.
Facts:
F1: Laura is female.
F2: Elias is male.
F3: Fabian is male.
F4: Claudia is female.
F5: Elena is female.
F6: Thomas is male.
F7: Amelie is female.
F8: Luisa is female.
F9: Patrick is male.
F10: Emilia is female.
F11: Samuel is male.
F12: Alina is female.
F13: Jonathan is male.
F14: Philipp is male.
F15: Nico is male.
F16: David is male.
F17: Emily is female.
F18: Konstantin is male.
F19: Florian is male.
F20: Helga is female.
F21: Nina is female.
F22: Lea is female.
F23: Felix is male.
F24: Leonie is female.
F25: Stefan is male.
F26: Gabriel is male.
F27: Tobias is male.
F28: Laura is parent of Fabian.
F29: Laura is parent of Felix.
F30: Laura is parent of Claudia.
F31: Elias is parent of Fabian.
F32: Elias is parent of Felix.
F33: Elias is parent of Claudia.
F34: Alina is parent of David.
F35: Alina is parent of Lea.
F36: Nico is parent of David.
F37: Nico is parent of Lea.
F38: Emily is parent of Nico.
F39: Konstantin is parent of Nico.
F40: Fabian is parent of Thomas.
F41: Fabian is parent of Amelie.
F42: Nina is parent of Tobias.
F43: Leonie is parent of Emily.
F44: Stefan is parent of Emily.
F45: Gabriel is parent of Tobias.
F46: Elena is parent of Thomas.
F47: Elena is parent of Amelie.
F48: Thomas is parent of Helga.
F49: Thomas is parent of Nina.
F50: Thomas is parent of Patrick.
F51: Luisa is parent of Helga.
F52: Luisa is parent of Nina.
F53: Luisa is parent of Patrick.
F54: Patrick is parent of Samuel.
F55: Patrick is parent of Alina.
F56: Patrick is parent of Jonathan.
F57: Patrick is parent of Philipp.
F58: Patrick is parent of Florian.
F59: Emilia is parent of Samuel.
F60: Emilia is parent of Alina.
F61: Emilia is parent of Jonathan.
F62: Emilia is parent of Philipp.
F63: Emilia is parent of Florian.

Unknown fact: Jonathan is aunt of Thomas.
The answer (True or False) is: 
\end{lstlisting}

\section{Examples of ProofWriter}
In this section, we provide examples of deduction experiments conducted on the ProofWriter Depth-1 dataset. We present examples for both the \textit{Semantics} and \textit{Symbols} settings.
\subsection{Semantics}
\begin{lstlisting}
The bear likes the dog.
The cow is round.
The cow likes the bear.
The cow needs the bear.
The dog needs the squirrel.
The dog sees the cow.
The squirrel needs the dog.
If someone is round then they like the squirrel.
If the bear is round and the bear likes the squirrel then the squirrel needs the bear.
If the cow needs the dog then the cow is cold.
Does it imply that the statement "The cow likes the squirrel." is True?
\end{lstlisting}

\begin{lstlisting}
The bear likes the dog.
The cow is round.
The cow likes the bear.
The cow needs the bear.
The dog needs the squirrel.
The dog sees the cow.
The squirrel needs the dog.
If someone is round then they like the squirrel.
If the bear is round and the bear likes the squirrel then the squirrel needs the bear.
If the cow needs the dog then the cow is cold.
Does it imply that the statement "The cow does not like the squirrel." is True?
\end{lstlisting}

\begin{lstlisting}
The bear likes the dog.
The cow is round.
The cow likes the bear.
The cow needs the bear.
The dog needs the squirrel.
The dog sees the cow.
The squirrel needs the dog.
If someone is round then they like the squirrel.
If the bear is round and the bear likes the squirrel then the squirrel needs the bear.
If the cow needs the dog then the cow is cold.
Does it imply that the statement "The cow likes the squirrel." is True?
\end{lstlisting}

\subsection{Symbols}
\begin{lstlisting}
The e4 likes the e5.
The e14 is e2.
The e14 likes the e4.
The e14 needs the e4.
The e5 needs the e26.
The e5 sees the e14.
The e26 needs the e5.
If someone is e2 then they like the e26.
If the e4 is e2 and the e4 likes the e26 then the e26 needs the e4.
If the e14 needs the e5 then the e14 is e1.
Does it imply that the statement "The e14 likes the e26." is True?
\end{lstlisting}

\begin{lstlisting}
The e27 is e7.
The e27 is e15.
The e30 does not chase the e27.
The e30 eats the e27.
The e30 is e1.
The e30 is e15.
The e30 visits the e27.
If something visits the e27 then the e27 does not visit the e30.
If something is e1 and e15 then it visits the e30.
Does it imply that the statement "The e30 visits the e30." is True?
\end{lstlisting}

\begin{lstlisting}
The e27 is e7.
The e27 is e15.
The e30 does not chase the e27.
The e30 eats the e27.
The e30 is e1.
The e30 is e15.
The e30 visits the e27.
If something visits the e27 then the e27 does not visit the e30.
If something is e1 and e15 then it visits the e30.
Does it imply that the statement "The e30 visits the e30." is True?
\end{lstlisting}

\section{Different Zero-Shot prompting}
We try different Zero-Shot prompts:

(1)
\begin{lstlisting}
I will provide a set of logical rules L1 to L{number of rules} and facts F1 to F{number of basic facts}. Please select one single logical rule from L1 to L{number of rules} and a few facts from F1 to F{number of basic facts} to predict True/False of the unknown fact using deductive reasoning.
Logical rules: {rules}
Facts: {basic facts}
Unknown fact: {unknown fact}
The answer (True or False) is:
\end{lstlisting}

(2) 
\begin{lstlisting}
I will provide a set of logical rules L1 to L{number of rules} and facts F1 to F{number of basic facts}. Please predict True/False of the unknown fact using deductive reasoning.
Logical rules: {rules}
Facts: {basic facts}
Unknown fact: {unknown fact}
The answer (True or False) is:
    
\end{lstlisting}

(3) 
\begin{lstlisting}
Given a set of rules and facts, you have to reason whether a statement is True or False.
Here are some rules: {rules}
Here are some facts: {basic facts}
Does it imply that the statement "{unknown fact}" is True?
The answer (YES or NO) is:
\end{lstlisting}

The results of the three prompts in the Zero-Shot setting are presented in Table~\ref{tab:zero-shot}. Among the three prompts, we select the one that achieves the best performance as our Zero-Shot prompt.

\begin{table}[h]
    \centering
    \begin{tabular}{c|ccc }
    \toprule
    & prompt1 & prompt2 & prompt3 \\
    \midrule
    $\text{KG}_1$ &54.5 &51.5 &53.8 \\ 
    
    \toprule
    \end{tabular}
    \caption{Different Zero-Shot Prompts of deductive reasoning. Results are in \%.}
    \label{tab:zero-shot}
\end{table}


\section{Introduction of Neo4j}\label{app:neo4j}
The Symbolic Tree is also a knowledge graph dataset. We conduct a comparison between the memorization abilities of a popular graph database, \textbf{Neo4j}, and \acp{llm}, \textbf{LLaMA-7B}. Neo4j is a widely used graph database system that provides convenient operations such as querying, inserting, deleting, and revising knowledge graphs. For our comparison, we deployed Neo4j on a high-performance server equipped with 2 Intel(R) Xeon(R) Platinum 8380 CPUs, each with 40 cores and 80 threads. The server has 512GB of memory and 4x1.8T NVME SSD disks. 

To ensure a fair comparison, we configured Neo4j with a pre-stored knowledge base that has a comparable disk space size to the LLaMA language model. Specifically, we used the Freebase dataset for Neo4j, which occupies approximately 30GB of disk space after preprocessing. For the language model, we used LLaMA-7B, which requires about 14GB of disk space. By comparing the performance of Neo4j and LLaMA-7B in terms of their memorization abilities, we can gain insights into the advantages and limitations of graph databases and language models for storing and retrieving knowledge.


\section{Task definitions}\label{app:evaluation of induction}

We define a few tasks to evaluate \acp{llm}' abilities of three kinds of reasoning and memorization.

\begin{itemize}[leftmargin=.2in, topsep=0pt]

\item \textit{deductive reasoning:} we use \textit{hypothesis classification}, \ie, predict the \textit{correctness} of the \textit{hypothesis} given the \textit{theory} where \textit{theory} consists of basic facts and logical rules, \textit{correctness} can be true or false, and \textit{hypothesis} is a predicted fact, which is one of the inferred facts or negative samples. The accuracy is the proportion of correct predictions.

\item \textit{inductive reasoning:}\label{inductive_reasoning} we perform the \textit{rule generation} task. Given multiple facts with similar patterns and a rule template, the goal is to induce a rule that entails these facts. Specifically, for each relation $r$, we use basic facts and those inferred facts that contain only relation $r$ as provided facts. The induced rule is generated after filling in the rule template. We test the generated rules against the ground truth rules. If the generated rule matches the ground truth rule exactly, we predict the rule to be correct; otherwise, we predict the rule to be incorrect. The precision is the proportion of correct predictions. Note that considering logical rules maybe not all chain rules (e.g., $r_1(y,x) \land r_2(y,z) \rightarrow r_3(x,z)$), we add inverse relation for each relation in order to transform them into chain rules and simplify the rule template (e.g., $r_1^{-1}(x,y) \land r_2(y,z) \rightarrow r_3(x,z)$). Furthermore, we provide a rule template for each relation. Take $auntOf$ as example, its rule template can be $\forall x,y,z: \text{\#\#}(x,y) \land\text{\#\#}(y,z) \land \text{++}(x) \rightarrow auntOf(x,z) $ or ``If x is \text{\#\#} of y and y is \text{\#\#} of z and x is ++, then x is aunt of z.", where \text{\#\#} can be $parent$ or $inverse\_parent$, ++ can be $female$ or $male$.

Besides, a single rule can be equivalent to multiple rules. For example, the rule $\forall x,y,z:  \text{parentOf}(x, y) \land \text{parentOf}(y, z) \land \text{gender}(x, \text{female}) \rightarrow \text{GrandmotherOf}(x,z) $ can be represented as $ \forall x,y,z:  \text{parentOf}(x, y) \land \text{parentOf}(y, z) \rightarrow \text{GrandparentOf}(x,z), \text{GrandparentOf}(x,z) \land \text{gender}(x, \text{female}) \rightarrow \text{GrandmotherOf}(x,z)$. We conduct the experiments with both rule representations and find single-longer rules perform better than multiple-short rules. Results are presented in Appendix~\ref{short rules}. Based on these observations and considering the simplicity of induction evaluation, we rewrite all logical rules by including only the $parentOf$ and $gender$ relations in the rule body. This also ensures that each inferred relation is implied by a single logical rule, referred to as \textit{grounding truth rule}.


\item \textit{abductive reasoning:} We use \textit{explanation generation} to evaluate abductive reasoning abilities. Given a \textit{theory} including basic facts and all logical rules, the task is to select specific facts and a logical rule to explain the \textit{observation}. The \textit{observation} is chosen from inferred facts. We use Proof Accuracy (PA) as an evaluation metric, i.e., the fraction of examples where the generated proof matches exactly any of the gold proofs. 

\item \textit{memorization:}
We use a subset of Symbolic Trees to fine-tune  the language model. For the symbolic setting, we use $r_1, r_2, r_3$ to replace the original relations in the semantic setting. Note that the new dataset does not overlap with the old knowledge base of LLMs, ensuring no disambiguation problem and the influence of pre-existing knowledge.
When memorizing, we use \textit{time}, \textit{efficiency} and \textit{forgetting} as metrics: \textit{time} is the cost time of adding/updating facts, \textit{efficiency} is the MRR (mean reciprocal rank~\cite{bordes2013translating}) of facts added/updated, and \textit{forgetting} is the MRR of the facts that should not be updated. When evaluating whether a fact has been successfully added or updated, we query LLM with a question about the tail entity and rank the probability of all tokens between all tail entities. The better LLM remembers a triplet, the higher the MRR gets. Note that, there may be more than one entity for each (head, relation) pair. We only consider the rank one of them.


\end{itemize}

\section{Implementation of memorization}\label{implementation of memorization}

We selected 1258 triplets from 4 Symbolic Trees to evaluate the effectiveness of adding knowledge. Following the prompting of~\citet{alpaca}, we use the head entity and relation as instructions and provide all candidate tails as input. The model's training objective is to autoregress toward the true tail entities. The detailed prompting is contained in Appendix~\ref{finetune_prompt}. In the updating step, we fine-tune the model on all 620 triplets from the first two trees whose tail entities are randomly flipped to false ones. Besides the effectiveness of updating, we evaluated the forgetting ratio using the remaining 638 triplets of the least two trees. These triplets have been remembered in the first step and haven't been updated in the second. Noting that, within each tree, the relationships between entities are independent, and the entities are distinct. Therefore, we propose that LLM should retain its memory of the previously remembered triplets when updating based on the first two trees. We utilized 4 A100 80G GPUs with batch size 64 for finetuning. The training process involved 100 epochs, employing a cosine learning rate schedule with an initial learning rate of 2e-5. We run these experiments three times and recorded their mean and standard MRR.

\section{Consistency of knowledge base}\label{updating}
In the context of updating a knowledge base, it is important to ensure the consistency of the knowledge base. When revising a fact, it is necessary to update other related facts accordingly to maintain a coherent and accurate knowledge base. For example, if we have the facts: Alice is Bob's mother, Amy is Alice's sister, Momo is Lily's sister, and Amy is Bob's aunt, and we update Bob's mother as Lily, it follows that Bob's aunt should be updated to Momo instead of Amy to maintain consistency.

However, in our experiments, we did not consider the situation of updating related facts in order to simplify the evaluation process, although it may lead to inconsistencies within the knowledge base. In real-world scenarios, ensuring consistency is an essential aspect of maintaining the accuracy and reliability of a knowledge base when updating or revising facts.


\section{Different Few-Shot-CoT examples}
We use three different kinds of demonstrations generation:

(1) leverage LLMs with the ``Let’s think step by step” prompt to generate reasoning chains for demonstrations.

(2) manually designed each demonstration with a question and a reasoning chain. A reasoning chain is composed of a rationale (a series of intermediate reasoning steps) and an expected answer.

(3) leverage LLMs with the ``Let’s think step by step” prompt to generate reasoning chains and filter with those generated chains coming with mistakes for demonstrations.

The results of these approaches are provided in Table~\ref{tab:different-few-shot-cot}. From the results, we observe that the \textit{Auto-Randomly} approach outperforms both the \textit{Auto-Filter} and \textit{Manually-Designed} approaches. This suggests that randomly generating reasoning chains is more suitable for \acp{llm} and can generate a diverse range of solutions, resulting in better reasoning performance.

\subsection{Few-Shot-CoT examples with Auto-Filter}
Statement: $r8$(Elena, Nina) \\
Answer: We can use logical rule L5: $\forall A,B,C: r3(A, B) \land r3(B, C) \land r2(A) \rightarrow r8(A, C)$ to deduce whether the statement $r8$(Elena, Nina) is true or false. 
From the facts, we know that $r3$(Elena, Thomas) and $r3$(Thomas, Nina). Therefore, we can use L5 with $A=$Elena, $B=$Thomas, and $C=$Nina to see if $r8$(Elena, Nina) is true or false. 
L5 states that if $r3$(A, B) and $r3$(B, C) and $r2$(A) are true, then $r8$(A, C) is true. In this case, we have $r3$(Elena, Thomas) and $r3$(Thomas, Nina) as true facts, and we also know that $r2$(Elena) is true from F5. Therefore, we can apply L5 to get:\\
$r8$(Elena, Nina) is true.\\
Therefore, the answer is True.

Statement: $r29$(Philipp, Elena)\\
Answer: We can use logical rule L26: $\forall A,B,C,D: r3(B, A) \land r3(C, B) \land r3(D, C) \land r1(A) \rightarrow r29(A, D)$ to infer that if $r3(B, A)$, $r3(C, B)$, $r3(D, C)$, and $r1(A)$ are true, then $r29(A, D)$ is also true. \\
From the given facts, we know that $r3$(Elena, Thomas) (F46) and $r3$(Thomas, Amelie) (F47). Therefore, we can use L26 with $A=$ Philipp, $B=$ Thomas, $C=$ Amelie, and $D=$ Elena to get:\\
$r3$(Thomas, Philipp) (from F62 and $r3(B, A)$) \\
$r3$(Amelie, Thomas) (from F47 and $r3(C, B)$) \\
$r3$(Elena, Amelie) (from F47 and $r3(D, C)$)\\ 
$r1$(Philipp) (from F14 and $r1(A)$) \\
Therefore, we can conclude that $r29$(Philipp, Elena) is True.

Statement: $r6$(Luisa, Helga)\\
Answer: We can use logical rule L3: $\forall A,B: r3(A, B) \land r2(A) \rightarrow r6(A, B)$ to deduce the truth value of the statement $r6$(Luisa, Helga). \\
From the facts, we know that $r3$(Luisa, Helga) and $r2$(Luisa) are true. Therefore, we can apply L3 to conclude that $r6$(Luisa, Helga) is true. \\
Thus, the answer is True.

Statement: $r14$(Amelie, Alina)\\
Answer: We can use logical rule L11: $\forall A,B,C,D,E: r3(B, A) \land r3(B, C) \land r3(C, D) \land r3(D, E) \land r2(A) \rightarrow r14(A, E)$ to infer $r14$(Amelie, Alina) if we can find facts that satisfy the antecedent of the rule. \\
From the facts, we can see that $r3$(Amelie, Thomas) and $r3$(Thomas, Helga) and $r3$(Helga, David) and $r3$(David, Alina) are true. Therefore, we can apply L11 with $A=$Amelie, $B=$Thomas, $C=$Helga, $D=$David, and $E=$Alina to get $r14$(Amelie, Alina) is True. \\
Therefore, the answer is True.

Statement: $r26$(Alina, Luisa)\\
Answer: We can use logical rule L23: $\forall A,B,C: r3(B, A) \land r3(C, B) \land r2(A) \rightarrow r26(A, C)$ to determine the truth value of the statement $r26$(Alina, Luisa). \\
From the facts, we know that $r3$(Alina, David) and $r3$(David, Luisa) (F34 and F52). Therefore, we can substitute A = Alina, B = David, and C = Luisa into L23 to get:\\
$r3$(David, Alina) $\land$ $r3$(Luisa, David) $\land$ $r2$(Alina) $\rightarrow$ $r26$(Alina, Luisa)\\
$r3$(Alina, David) is true (F34) and $r2$(Alina) is also true (F12). Therefore, we can simplify the above expression to:\\
$r3$(Luisa, David) $\rightarrow$ $r26$(Alina, Luisa)\\
$r3$(Luisa, David) is true (F52), so we can conclude that $r26$(Alina, Luisa) is true. \\
Therefore, the answer is True.

\subsection{Few-Shot-CoT examples with Manually-Designed}
Statement: $r14$(Amelie, Jonathan)\\
Answer: We can use logical rule L11: $\forall A,B,C,D,E: r3(B, A) \land r3(B, C) \land r3(C, D) \land r3(D, E) \land r2(A) \rightarrow r14(A, E)$. We can substitute A = Amelie and E = Jonathan to get: $r3 (B, Amelie) \land r3 (B, C) \land r3 (C, D) \land r3 (D, Jonathan) \land r2 (Amelie) \rightarrow r14 (Amelie, Jonathan)$. Now, we need to find all facts that match the conditions $r3$(B, Amelie), $r3$(B, C), $r3$(C, D), $r3$(D, Jonathan) and $r2$(Amelie). We have:\\
F41: $r3$(Fabian, Amelie)\\
F47: $r3$(Elena, Amelie)\\
F56: $r3$(Patrick, Jonathan)\\
F61: $r3$(Emilia, Jonathan)\\
F7: $r2$(Amelie)\\
We can see that F41 and F47 match the condition $r3$(B, Amelie) with $B$ being Fabian and Elena, respectively. F56 and F61 match the condition $r3$(D, Jonathan) with $D$ being Patrick and Emilia, respectively. F7 matches the condition $r2$(Amelie). Therefore, we have two possible ways of assigning values to B, C, and D:\\
(1) B = Fabian, D = Patrick.\\
(2) B = Elena, D = Emilia.\\
(3) B = Fabian, D = Emilia.\\
(4) B = Elena, D = Patrick.\\
Using the values of B and D from the possible assignments above, we can find the value of C that satisfies the conditions $r3$(B, C) and $r3$(C, D). For assignment (1), we have:\\
F40: $r3$(Fabian, Thomas)\\
F50: $r3$(Thomas, Patrick)\\
Since all the conditions are satisfied, we can conclude that the statement $r14$(Amelie, Jonathan) is True if we assign B = Fabian and C = Thomas and D = Patrick. Thus, the output is True.

Statement: $r31$(Philipp, Nina)\\
Answer: We can use logical rule L28: $\forall A,B,C,D: r3(B, A) \land r3(C, B) \land r3(C, D) \land r1(A) \rightarrow r31(A, D)$. We can substitute A = Philipp and D = Nina to get: $r3 (B, Philipp) \land r3 (C, B) \land r3 (C, Nina) \land r1 (Philipp) \rightarrow r31 (Philipp, Nina)$. Now, we need to find all facts that match the conditions $r3$(B, Philipp), $r3$(C, B), $r3$(C, Nina), and $r1$(Philipp). We have:\\
F57: $r3$(Patrick, Philipp)\\
F62: $r3$(Emilia, Philipp)\\
F49: $r3$(Thomas, Nina)\\
F52: $r3$(Luisa, Nina)\\
We can see that F57 and F62 match the condition $r3$(B, Philipp) with $B$ being Patrick and Emilia, respectively. F49 and F52 match the condition $r3$(C, Nina) with $C$ being Thomas and Luisa, respectively. Therefore, there is only one single way of assigning values to B and C.\\
B = Patrick and C = Thomas. We can substitute B and C to get: $r3 (Patrick, Philipp) \land r3 (Thomas, Patrick) \land r3 (Thomas, Nina) \land r1 (Philipp) \rightarrow r31 (Philipp, Nina)$. Since all the conditions are satisfied, we can conclude that the statement $r31$ (Philipp, Nina) is True if we assign B = Patrick and C = Thomas. Thus, the output is True.

Statement: $r8$(Leonie, Nico)\\
Answer: We can use logical rule L5: $\forall A,B,C: r3(A, B) \land r3(B, C) \land r2(A) \rightarrow r8(A, C)$. We can substitute A = Leonie and C = Nico to get: $r3 (Leonie, B) \land r3 (B, Nico) \land r2 (Leonie) \rightarrow r8 (Leonie, Nico)$. Now, we need to find all facts that match the conditions $r3$(Leonie, B), $r3$(B, Nico), and $r2$(Leonie). We have:\\
F43: $r3$(Leonie, Emily)\\
F38: $r3$(Emily, Nico)\\
F39: $r3$(Konstantin, Nico)\\
F24: $r2$(Leonie)\\
We can see that F43 matches the condition $r3$(Leonie, B) with $B$ being Emily. and F38 matches the condition $r3$(B, Nico) with $B$ being Emily. Therefore, we can assign B = Emily. Also, F24 matches the condition $r2$(Leonie). Therefore, we can conclude that the statement $r8$(Leonie, Nico) is True if we assign B = Emily. Therefore, the output is True.

Statement: $r7$(Patrick, Alina)\\
Answer: We can use logical rule L4: $\forall A,B: r3(A, B) \land r1(A) \rightarrow r7(A, B)$. We can substitute A = Patrick and B = Alina to get: $r3 (Patrick, Alina) \land r1 (Patrick) \rightarrow r7 (Patrick, Alina)$. Now, we need to find all facts that match the conditions $r3$(Patrick, Alina) and $r1$(Patrick). We have:\\
F55: $r3$(Patrick, Alina)\\
F9: $r1$(Patrick)\\
We can see that F55 matches the condition $r3$(Patrick, Alina), and F9 matches the condition $r1$(Patrick). Therefore, we can conclude that the statement $r7$(Patrick, Alina) is True based on L4. Thus, the output is True.

\begin{table}[!t]\small

  \caption{The deductive reasoning results of one Symbolic Tree KG with ChatGPT. Results are in \%.}
  \label{tab:different-few-shot-cot}
  \centering
  \resizebox{0.6\linewidth}{!}{
  \begin{tabular}{c|ccc}
    \toprule

    & Auto-Randomly & Auto-Filter & Manually-Designed \\
    \midrule
    $\text{KG}_1$ &55.6  &53.6 &40.7 \\ 
    \toprule
  \end{tabular}
  }
  
\end{table}

\section{Reasoning results of after human selection}\label{app:selection}
In this section, we perform all experiments after human selection, providing the \acp{llm} with only the relevant facts related to the predicted fact.

For example, we have facts: \\
F1: Lea is female.\\
F2: Philipp is male.\\
F3: Leonie is female.\\
F4: Patrick is parent of Alina.\\
F5: Patrick is parent of Philipp.\\
F6: Nico is parent of David.\\
F7: Nico is parent of Lea.\\
F8: Laura is parent of Fabian.\\
F9: Laura is parent of Felix.

If we predict the unknown fact "Philipp is the brother of Alina", we filter out the irrelevant facts and retain only the related facts:\\
F1: Philipp is male.\\
F2: Patrick is parent of Alina.\\
F3: Patrick is parent of Philipp.\\

\subsection{Comparison of Removing rules and Semantics}\label{app:selection-deduction}
To explore the ability of \acp{llm} to utilize internal and external knowledge, we conduct an additional experiment where we provided \acp{llm} with only the relevant facts related to the predicted fact. We compared the performance of \textit{Removing rules} (leveraging internal knowledge) with \textit{Semantics} (providing external logical rules). Surprisingly, from Table~\ref{tab:comparison of rev and sem}. we find that \textit{Removing rules} performed better than \textit{Semantics}. This suggests that \acp{llm} possess the necessary internal knowledge to support answering questions and reasoning tasks, and leveraging this internal knowledge is more effective for reasoning than relying on external logical rules.

\begin{table}[h]
    \centering
    \begin{tabular}{c|cc}
    \toprule
        &Removing rules &Semantics  \\
        \midrule
         $\text{KG}_1$ &78.2 &72.7 \\
    \toprule
    \end{tabular}
    \caption{Comparison of Removing rules and Semantics of deductive reasoning. Results are in \%.}
    \label{tab:comparison of rev and sem}
\end{table}

\acp{llm}' output examples are provided below:

\textbf{Semantics}:
According to the given logical rule, for A to be a granddaughter of D, B must be the parent of A and C must be the parent of B. However, there is no direct connection between Emilia and Lea in the given facts. Although Emilia is the parent of Alina, who is the parent of Lea, Emilia is not the parent of Lea's parent. Therefore, we cannot conclude that Lea is Emilia's granddaughter based on the given facts and logical rule.

\textbf{Removing rules}:
Based on the given facts, Emilia is the parent of Alina, who is the parent of Lea. Therefore, Emilia is the grandmother of Lea, making Lea Emilia's granddaughter.

From the example, we can observe that when relying on external logical rules, \acp{llm} need to strictly adhere to the reasoning process, which can be more challenging for \acp{llm} to predict unknown answers compared to utilizing the commonsense knowledge already contained within \acp{llm}. This suggests that leveraging the internal knowledge of \acp{llm} can be more effective for reasoning tasks.


\subsection{Reasoning results after human selection}\label{app:selection-induction}
We conduct deductive and inductive reasoning experiments to examine the performance of \acp{llm} when only provided with the relevant facts related to the predicted fact. The results are presented in Table~\ref{tab:after-selection}. They demonstrate that after selecting useful information, \acp{llm} perform reasoning tasks more effectively. This finding suggests that \acp{llm} face challenges when processing excessively long in-context information. Selecting relevant facts helps to reduce the memorization load on \acp{llm} and enables them to focus on the most relevant information for reasoning, leading to improved performance.


\begin{table}[h]
    \centering
    \begin{tabular}{cc|ccc}
    \toprule
          & &Zero-Shot  &Zero-Shot-CoT  \\
          \midrule
          \multirow{2}{*}{\textbf{\shortstack{Deductive}}} & standard &52.6 &56.1  \\
          & after-selection &50.8 &59.5  \\
         \midrule
         \multirow{2}{*}{\textbf{\shortstack{Inductive}}}   & standard  &7.14 &7.14 \\
          & after-selection &67.9 &67.9 \\
    \toprule
    \end{tabular}
    \caption{ Reasoning results after human selection. Results are \%. }
    \label{tab:after-selection}
\end{table}

\section{Reasoning with natural language}
In this section, we conducted experiments using the \textit{Symbols} setting with deduction, induction, and abduction on a Symbolic Tree dataset expressed in natural language. The results are presented in Table~\ref{tab:natural language results}. We observed that, in general, \acp{llm} performed better when using logical language compared to natural language.
\begin{table}\small
  \caption{The reasoning results of Symbolic Tree (expressed in natural language). Results are in \%.}
  \label{tab:natural language results}
  \centering
  \begin{tabular}{ccc|ccc}
    \toprule
    \textbf{Category} &
    \textbf{Model}  &
    \textbf{Baseline} & deduction & induction & abduction \\
    \midrule

    \multirow{3}{*}{\textbf{\shortstack{Logic language}}} 
    & \multirow{3}{*}
    {\textbf{\shortstack{ChatGPT}}} 
      & Zero-Shot &52.6 &7.14 &1.95     \\
   
   & & Zero-Shot-CoT  &56.1 &7.14 &3.57      \\
    & & Few-Shot-CoT    &53.7 &- &13.3 \\

    \midrule
    
    \multirow{3}{*}{\textbf{\shortstack{Natural language}}} 
    & \multirow{3}{*}
    {\textbf{\shortstack{ChatGPT}}} 
      & Zero-Shot     &50.6 &3.57 &3.90 \\
   
   & & Zero-Shot-CoT  &50.2 &7.14 &1.95      \\
    & & Few-Shot-CoT  &51.9 &- &8.13 \\

    \toprule

  \end{tabular}
\end{table}

\section{Reasoning results of two representations  }\label{app:results of two repre}
For the Symbolic Tree dataset, facts and rules can be represented as logic language and natural language text as the input of \acp{llm}. For example, the fact ``motherOf(Alice, Bob)'' can be represented as ``Alice is Bob's mother''; the fact ``r1(Alice, Bob) can be represented as ``Alice is r1 of Bob''; the rule ``$\forall x,y: \text{parentOf}(x,y) \rightarrow \text{childOf}(y,x)$'' can be represented as ``If x is parent of y, then y is parent of x.''. Through numerous trials, we find that for the \textit{Symbols} or \textit{Counter-CS} setting, \acp{llm} tend to perform better when using logic language representations. Conversely, for the \textit{Semantics} setting, \acp{llm} tend to perform better when using natural language text. The results are presented in Table~\ref{tab:two repre}. These observations suggest that natural language representations better stimulate the semantic understanding capabilities of \acp{llm}, while logical language representations are more conducive to symbolic reasoning.

\begin{table}[h]
    \centering
    \begin{tabular}{cc|ccc}
    \toprule
          & &Zero-Shot  &Zero-Shot-CoT  \\
          \midrule
          \multirow{2}{*}{\textbf{\shortstack{Symbols}}} & logic &52.6 &56.1  \\
          & natural language &49.0 &51.1 \\
         \midrule
         \multirow{2}{*}{\textbf{\shortstack{Semantics}}} & logic &61.4 &61.9 \\
          & natural language &69.3 &64.3\\
         \midrule
         \multirow{2}{*}{\textbf{\shortstack{Counter-CS}}} & logic &52.6 &54.4  \\
          & natural language &48.7 &48.3 \\

    \toprule
    \end{tabular}
    \caption{ Deductive reasoning results of one Symbolic Tree in different representations. Results are \%.}
    \label{tab:two repre}
\end{table}

\section{Reasoning with garbled symbols}\label{garbled symbols}
In this section, we randomly sample 4-8 letters to construct a garbled symbols word as each relation label. However, because LLMs process text by tokens, common sequences of characters found in the text, this random label is often tokenized as a single letter by \acp{llm}. This leads to difficulty for \acp{llm} to comprehend the word-level meaning of the label, harming the performance of reasoning. We also attempted to use single-token garbled symbols as relation labels, but similar challenges were encountered, with the labels often being interpreted as typos by \acp{llm}, further hindering their reasoning abilities. Our experiments and case studies also demonstrate this point.

As shown in Table~\ref{garbled}, using ID symbols (e.g., ``r1'') as relation labels performs better than using garbled symbols or single-token symbols.

Some examples with garbled symbols:

Statement: $ckxovsaq$(Alina, Philipp)\\
Answer: We can use logical rule L1 to make a deduction about Alina and Philipp. L1 states that if there is a person B who is a friend of both A and C, and A is not \underline{outgoing}, then A \underline{trusts} C. We know that Alina is not outgoing (F12: $ntoea$(Alina)) and that Philipp is a friend of Emilia (F62: $yufevh$(Emilia, Philipp)). However, we do not have any information about Alina and Emilia's friendship. Therefore, we cannot use L1 to make a deduction about $ckxovsaq$(Alina, Philipp). The answer is False.

Statement: $nyheg$(Emily, Nico) \\
Answer: We can use logical rule L3: $\forall A,B: yufevh(A,B) \land ntoea(A) \rightarrow nyheg(A,B)$ to determine if the statement is true or false. This rule states that if A is not outgoing and there is a friendship relation between A and B, then A is \underline{jealous} of B. 
From the facts, we know that Emily is not outgoing (F17: $ntoea$(Emily)) and she is friends with Nico (F38: $yufevh$(Emily, Nico)). Therefore, we can apply L3 to conclude that $nyheg$(Emily, Nico) is true. 
Therefore, the answer is True.

Statement: $phnd$(Luisa, Jonathan)\\
Answer: We can use logical rule L5: $\forall A,B,C: yufevh(A,B) \land yufevh(B,C) \land ntoea(A) \rightarrow phnd(A,C)$ to determine if the statement is true or false. This rule states that if A is not outgoing and there is a path from A to C through B, then A is \underline{phoned} by C. 
From the given facts, we know that Luisa is not outgoing (F8: $ntoea$(Luisa)). We also know that there is a path from Luisa to Jonathan through Patrick (F53: $yufevh$(Luisa, Patrick) and F56: $yufevh$(Patrick, Jonathan)). Therefore, we can apply L5 with A = Luisa, B = Patrick, and C = Jonathan to get:
$phnd$(Luisa, Jonathan) is true.
Therefore, the answer is True.

\begin{table}[!t]\small

  \caption{The deductive reasoning results of one Symbolic Tree KG with ChatGPT. Results are in \%.}
  \label{garbled}
  \centering
  \resizebox{\linewidth}{!}{
  \begin{tabular}{c|cccc}
    \toprule

    & Zero-Shot & Zero-Shot-CoT & Few-Shot-CoT & Zero-Plus-Few-Shot-CoT  \\
    \midrule
   
    garbled symbols &52.4 &54.4 &55.5 &52.9  \\

    single-token symbols &52.5 &53.2 &53.9 &52.3 \\
     ID symbols &54.5 &57.0 &57.2 &52.3 \\
    
    \midrule
    short rules  &52.1 &51.1 &52.9 &54.6\\
     long rules &54.5 &57.0 &57.2 &52.3\\

    \toprule
  \end{tabular}
  }
  
\end{table}

Some examples with single-token symbols:

Statement: $iance$(Patrick, Lea)\\
Answer: We can use logical rule L4: $\forall A,B: inely(A, B) \land atis(A) \rightarrow lesai(A, B)$, which states that if A is an \underline{artist} and B is in A's \underline{inner circle}, then B is a \underline{muse} of A. We can also use fact F53: $inely$(Luisa, Patrick), which means that Luisa is in Patrick's inner circle. Therefore, we can conclude that $lesai$(Patrick, Luisa) is true. \\
Next, we can use logical rule L6: $\forall A,B,C: inely(A, B) \land inely(B, C) \land atis(A) \rightarrow iance(A, C)$, which states that if A is an artist, B is in A's inner circle, and C is in B's inner circle, then A and C have an \underline{artistic alliance}. We can use fact F35: $inely$(Alina, Lea), which means that Alina is in Lea's inner circle. Therefore, we can conclude that $iance$(Alina, Lea) is true.\\
Finally, we can use logical rule L1: $\forall A,B,C: inely(B, A) \land inely(B, C) \land uitka(A) \rightarrow reib(A, C)$, which states that if A is a \underline{writer}, B is in A's inner circle, and B is also in C's inner circle, then A and C have a literary rivalry. We can use fact F54: $inely$(Patrick, Samuel), which means that Samuel is in Patrick's inner circle. We can also use fact F35: $inely$(Alina, Lea), which means that Alina is in Lea's inner circle. Therefore, we can conclude that $reib$(Patrick, Lea) is true.\\
Since Patrick has a \underline{literary rivalry} with Lea, but not an artistic alliance, we can conclude that the statement $iance$(Patrick, Lea) is False.

Statement: $icers$(Jonathan, David)\\
Answer: We can use logical rule L10: $\forall A,B,C,D: inely(B, A) \land inely(B, C) \land inely(C, D) \land atis(A) \rightarrow icers(A, D)$, which states that if A is an artist and B, C, and D are in a line with B in the middle, then A influences D. \\
We can use the following facts:
- $inely$(Jonathan, Alina) (F55)\\
- $inely$(Alina, David) (F34)\\
From these facts, we can conclude that Jonathan, Alina, and David are in a line with Alina in the middle. Also, we know that Alina is an artist. Therefore, we can apply L10 to conclude that Alina influences David. \\
Since Alina influences David, we can use L4: $\forall A,B: inely(A, B) \land atis(A) \rightarrow lesai(A, B)$, which states that if A influences B and A is an artist, then B is a musician. Therefore, we can conclude that David is a \underline{musician}. \\
Therefore, the statement $icers$(Jonathan, David) is False, since David is not an \underline{ice skater}.

Statement: $indr$(Florian, Nina)\\
Answer:  We can use logical rule L24: $\forall A,B: inely(B, A) \land atis(A) \rightarrow indr(A, B)$, which states that if A is an \underline{artist} and B is a \underline{fan} of A, then B is indirectly a fan of all the artists that A is a fan of. \\
From the given facts, we know that $atis$(Florian) and $inely$(Emilia, Nina). Therefore, we can conclude that Nina is indirectly a fan of Florian, since Emilia is a fan of Florian and Nina is a fan of Emilia. \\
Thus, the answer is True.

\section{More reasoning results of Symbolic Tree}
We experiment with 10 sampled trees and report the average results in the main body. 
In this section, we provide the reasoning results of each sampled Symbolic Tree, presented in Table~\Cref{tab:deductive,tab:inductive,tab:abductive}.

\begin{table}[t]
  \caption{The deductive reasoning results of Symbolic Tree datasets. Results are in \%.}
  \label{tab:deductive}
  \centering
  \resizebox{\linewidth}{!}{
  \begin{tabular}{ccc|ccccccccccc}
    \toprule
    \textbf{Category} &
    \textbf{Model}  &
    \textbf{Baseline}  &
    S$_1$ & S$_2$ & S$_3$ & S$_4$ & S$_5$ & S$_6$ & S$_7$ & S$_8$ & S$_9$ & S$_{10}$ & Avg.\\
    \midrule

    \multirow{5}{*}{\textbf{\shortstack{Symbols}}} &
    \multirow{1}{*}{\textbf{\shortstack{Random}}} &- & 52.4 &50.8 & 51.3 &50.2 &49.3 &49.1 &48.1 &52.3 &48.4 &49.0 &50.1\\
    & \multirow{4}{*}
    {\textbf{\shortstack{ChatGPT}}}
      & Zero-Shot    &52.6 &50.6 &50.5 &49.5 &55.2 &53.1 &50.0 &53.4 &56.6 &54.0 &52.6 \\

   & & Zero-Shot-CoT &56.1 &57.0 &55.4 &57.0 &54.5 &56.1 &55.5 &56.9 &50.0 &58.0   &55.7       \\
    & & Few-Shot-CoT &53.7 &56.9 &55.2 &54.4 &55.1 &52.0 &54.0 &55.8 &56.8 &54.5 &54.8 \\
    & & Zero-Plus-Few-Shot-CoT  &53.7 &53.6 &55.4 &51.4 &54.0 &50.9 &54.0  &54.2 &58.4 &54.5  &54.0 \\


\midrule
 \multirow{4}{*}{\textbf{\shortstack{Semantics}}} &
    \multirow{4}{*}
    {\textbf{\shortstack{ChatGPT}}}
      & Zero-Shot   &70.0 &64.8 &70.4 &65.8 &61.4 &63.8 &65.8 &67.4 &63.0 &68.9 &66.1\\
     & & Zero-Shot-CoT &66.7 &64.8 &64.6 &64.1 &64.4 &67.2 &66.5 &66.7 &64.6 &65.4 &65.5\\

   & & Few-Shot-CoT   &71.8 &70.4 &63.9 &69.2 &66.7 &59.3 &68.7 &68.3  &67.9 &64.4 &67.1  \\
    & & Zero-Plus-Few-Shot-CoT &71.3 &67.8 &66.6 &69.5  &65.7 &60.9 &68.4  &68.3 &66.5 &66.8 &67.2  \\


    \midrule
    & \textbf{Logic-based} &- &100 &100 &100 &100 &100 &100 &100 &100 &100 &100 &100  \\
    \toprule

  \end{tabular}
  }
\end{table}

\begin{table}[t]\small
  \caption{The inductive reasoning results of Symbolic Tree datasets. Results are in \%.}
  \label{tab:inductive}
  \centering
  \resizebox{\linewidth}{!}{
  \begin{tabular}{ccc|ccccccccccc}
    \toprule
    \textbf{Category} &
    \textbf{Model} &
    \textbf{Baseline} &
    S$_1$ & S$_2$ & S$_3$ & S$_4$ & S$_5$ & S$_6$ & S$_7$ & S$_8$ & S$_9$ & S$_{10}$ & Avg.\\

    \midrule

      \multirow{4}{*}{\textbf{\shortstack{Symbols}}}
      & \multirow{2}{*}{\textbf{\shortstack{ChatGPT}}}  & Zero-Shot &7.14 &9.09 &3.57  &7.14 &4.54 &14.3 &4.54 &7.14 &3.57 &0.0 & 6.10\\
   & & Zero-Shot-CoT &7.14 &7.14 &3.57 &14.3 &14.3 &7.14 &3.57  &0.0  &14.3 &7.14 &7.86 \\

    & \multirow{2}{*}{\textbf{\shortstack{GPT-4}}} & Zero-Shot & 14.3 &10.7  &10.7  &10.7  &7.14 &7.14 &10.7  &7.14 &7.14 &7.14 &9.28\\
   & & Zero-Shot-CoT &21.4 &7.14 &17.9 &7.14  &3.57  &7.14  &7.14  &7.14  &7.14  &3.57 &8.93\\

   \midrule

      \multirow{4}{*}{\textbf{\shortstack{Semantics}}}
      & \multirow{2}{*}{\textbf{\shortstack{ChatGPT}}}  & Zero-Shot &25.0 &32.1 &39.3 &39.3 &42.9 &39.3 &35.7 &32.1 &35.7 &42.9 &36.4 \\
   & & Zero-Shot-CoT &25.0 &28.6 &35.7 &28.6 &35.7 &35.7 &28.6 &35.7 &39.3 &28.6 &32.2 \\

    & \multirow{2}{*}{\textbf{\shortstack{GPT-4}}} & Zero-Shot &53.6  &53.6 &50.0 &53.6 &50.0 &53.6 &50.0 &57.1 &53.6 &50.0  &52.5 \\
   & & Zero-Shot-CoT &53.6 &57.1 &53.6 &53.6 &57.1 &53.6  &50.0  &53.6  &57.1  &50.0 &53.9\\


    \midrule
    & \textbf{Rule-based} &- &64.3 &60.7  &60.7 &46.4 &67.9 &50.0 &64.3 &57.1 &53.6 &46.4 &57.1 \\

    \toprule

  \end{tabular}
  }
\end{table}

\begin{table}[t]\small
  \caption{The abductive reasoning results of Symbolic Tree KGs. Results are in \%.}
  \label{tab:abductive}
  \centering
  \resizebox{\linewidth}{!}{
  \begin{tabular}{ccc|ccccccccccc}
    \toprule
    \textbf{Category} &
    \textbf{Model} &
    \textbf{Baseline} &
    S$_1$ & S$_2$ & S$_3$ & S$_4$ & S$_5$ & S$_6$ & S$_7$ & S$_8$ & S$_9$ & S$_{10}$ & Avg.\\
    \midrule

    \multirow{3}{*}{\textbf{\shortstack{Symbols}}} & \multirow{3}{*}{\textbf{\shortstack{ChatGPT}}}  & Zero-Shot &1.95 &0.31 &1.07 &1.52 &2.36 &1.45 &1.06 &0.75 &3.1 &1.39 &1.50\\
   & & Zero-Shot-CoT &3.57 &4.08 &5.00 &3.03  &3.70 &3.77 &5.28 &7.55 &7.78 &5.21 &4.90\\
   & & Few-Shot-CoT &13.3 &7.70 &8.39 &7.42 &10.8 &8.55 &10.7 &14.3 &8.95 &7.99 &9.81\\

    \midrule

    \multirow{3}{*}{\textbf{\shortstack{Semantics}}} & \multirow{3}{*}{\textbf{\shortstack{ChatGPT}}}  & Zero-Shot &1.95  &3.14  &3.57 &1.52 &2.69 &2.32 &3.87 &3.02 &3.89 &3.47 &2.94\\
   & & Zero-Shot-CoT  &4.22 &5.34 &4.64 &3.63 &2.69 &2.90 &4.23 &1.89 &3.11 &1.39 &3.40\\
    
   & & Few-Shot-CoT &9.90 &13.2 &10.9 &7.42 &8.59 &0.97 &11.3 & 13.0 & 11.3 &11.1 &9.77\\
    \midrule
     & \textbf{Rule-based} &- &100 &100 &100 &100 &100 &100 &100 &100 &100 &100 &100  \\
    \toprule

  \end{tabular}
  }

\end{table}

\section{Abductive reasoning on smaller datasets}\label{app:shorter context}
We use smaller Symbolic Tree datasets to conduct the abductive reasoning experiment, which contains about 12 entities and 100 facts. The results are provided in Table~\ref{tab:shorter context}. We compare \textit{Symbols} and \textit{Semantics} and find that the \textit{Semantics} setting still outperforms the \textit{Symbols} setting. This reinforces the hypothesis that preserving semantics enhances the reasoning capabilities of \acp{llm}. 

Additionally, abductive reasoning in a shorter context yielded better performance compared to a longer context. This suggests that the length of the context has an impact on reasoning performance. Shorter contexts make selecting relevant and useful information easier while minimizing the influence of unrelated content.

\begin{table}[h]\small
  \caption{The abductive reasoning results of symbolic datasets and Family-Tree datasets. Results are in \%.}
  \label{tab:shorter context}
  \centering
  \resizebox{0.6\linewidth}{!}{
  \begin{tabular}{cc|cc}
    \toprule
    \textbf{Category} &
    \textbf{Baseline} & 
    short context  & long context\\
    \midrule
    
    \multirow{2}{*}{\textbf{\shortstack{Symbols}}}

    & ChatGPT: Zero-Shot-CoT  &9.78 &3.57 \\
  & GPT-4: Zero-Shot-CoT &46.7 &32.1\\

\midrule
    \multirow{2}{*}{\textbf{\shortstack{Semantics}}} 
     
      & ChatGPT: Zero-Shot-CoT &5.43 &4.22 \\
  & GPT-4: Zero-Shot-CoT &59.8 &31.8\\

    \toprule

  \end{tabular}
  }
   
\end{table}

    



\section{More details on the experimental setup of commonsense reasoning}\label{GreaseLM}
CommonsenseQA is a multiple-choice question-answering dataset, including 12,102 questions with one correct answer and four distractor answers. OpenBookQA is also a question-answering dataset that requires combining an open book fact with broad common knowledge. It contains 5,957 multiple-choice elementary-level science questions (4,957 train, 500 dev, 500 test). For CommonsenseQA, we test on both validation and test set splits, while for OpenBookQA we only test on the test set. The task is that given a question and candidate answers in multiple choices, we evaluate the ability of commonsense reasoning by answering commonsense questions. The higher accuracy, the stronger the reasoning ability the model has. 

We use ChatGPT, GPT-4 and GreaseLM~\cite{zhang2022greaselm} as baselines. Experiments are conducted on few-shots and zero-shot prompts. Prompts used here are shown in Appendix~\ref{prompts}. GreaseLM has been validated to achieve significant performance on the two datasets mentioned above among similar works using KG.

While LLM can only provide textual knowledge limited to word-level embedding encoded from unstructured natural language, it falls short of interpretability and explainable predictions. Knowledge graphs (KG) in massive informative triplets usually play a critical role in providing complementary knowledge. It enables explicitly represented and learned structured data for better language understanding. Therefore, previous works demonstrate promising performance gains leveraging different LM architectures and KGs across different domains for improving commonsense reasoning. GreaseLM~\cite{zhang2022greaselm} enables fusion and exchange of information from both LM and KG across multiple layers and enables joint reasoning over both modalities. It has been validated to achieve significant performance on the two datasets mentioned above among similar works using KG. Here we leverage its results directly from the original paper for further comparison.

\section{Replacing entity labels}\label{entity labels}
In this section, we conducted experiments to investigate the effects of replacing entity names (such as ``Alice'') with entity IDs (\eg, ``e1'') in the context of reasoning tasks. The results are provided in Table~\ref{tab:replacing entities}. Comparing the performance of replacing relation names with replacing both entity and relation names, we observe that replacing entity names after replacing relation names had little impact on the overall performance.

Furthermore, we consider the scenario of only replacing entity names. Compared to the case of not replacing any labels, the results indicate that although replacing entity labels retains some level of semantics, it has a detrimental effect on reasoning performance. Additionally, we observed that the negative impact of decoupling the semantics of relations was more significant than that of decoupling the semantics of entities. These findings indicate a substantial portion of the semantic information is concentrated in the relation names.

\begin{table}[h]
    \centering
    \begin{tabular}{c|ccc}
    \toprule
     & Zero-Shot &Zero-Shot-CoT \\
    \midrule
    replacing none &69.3 &66.1 \\
    replacing ent &63.6 &58.9 \\
    replacing rel &54.5 &54.5\\
    replacing ent \& rel &57.5 &55.6 \\
    \toprule
    \end{tabular}
    \caption{Comparison of replacing entity labels in deductive reasoning experiment (ChatGPT). Results are in \%.}
    \label{tab:replacing entities}
    
\end{table}

\section{Multi-short rules}\label{short rules}
Besides, a single rule can be equivalent to multiple rules. For example, the rule $\forall x,y,z:  \text{parentOf}(x, y) \land \text{parentOf}(y, z) \land \text{gender}(x, \text{female}) \rightarrow \text{GrandmotherOf}(x,z) $ can be represented as $ \forall x,y,z:  \text{parentOf}(x, y) \land \text{parentOf}(y, z) \rightarrow \text{GrandparentOf}(x,z), \text{GrandparentOf}(x,z) \land \text{gender}(x, \text{female}) \rightarrow \text{GrandmotherOf}(x,z)$. We conduct the experiments with both rule representations and find single-longer rules perform better than multiple-short rules. Results are presented in Table~\ref{garbled}.

\end{appendices}
\clearpage




\end{document}